\documentclass{article}

\usepackage[preprint]{corl_2022} % Uncomment for pre-prints (e.g., arxiv); This is like ``final'', but will remove the CORL footnote.
\usepackage{graphicx, float, url, wrapfig}
\usepackage{amsmath, amssymb, bbm, mathtools}
\usepackage{bibunits}

\usepackage{algorithm}
\usepackage{algpseudocode}
\usepackage{booktabs}
\usepackage{multirow}
\usepackage{subfig}
\usepackage{enumitem}
\usepackage{amsmath,amsfonts,bm}

\def\eqref#1{equation~\ref{#1}}
\def\1{\bm{1}}

\def\vtheta{{\bm{\theta}}}

\DeclareMathAlphabet{\mathsfit}{\encodingdefault}{\sfdefault}{m}{sl}
\SetMathAlphabet{\mathsfit}{bold}{\encodingdefault}{\sfdefault}{bx}{n}

\providecommand{\ie}{\emph{i.e.,} }
\providecommand{\eg}{\emph{e.g.,} }
\providecommand{\parab}[1]{\noindent\textbf{#1}}

\usepackage{csquotes}

\usepackage[font=small]{caption}
\newcommand{\bftab}{\fontseries{b}\selectfont}
\usepackage{makecell, tabularx, multirow,  adjustbox}

\usepackage{longtable}
\usepackage{stackengine}
\newsavebox\mybox
\newcommand\Includegraphics[2][]{\sbox{\mybox}{%
  \includegraphics[#1]{#2}}\abovebaseline[-.5\ht\mybox]{%
  \addstackgap{\usebox{\mybox}}}}

\title{Leveraging Language for Accelerated Learning \\ of Tool Manipulation}

\author{
  Allen Z. Ren$^1$, Bharat Govil$^2$, Tsung-Yen Yang$^2$, Karthik Narasimhan$^2$\thanks{Equal contribution in advising}
, Anirudha Majumdar$^1$\footnotemark[1] \\
  $^1$Department of Mechanical and Aerospace Engineering\\
  $^2$Department of Computer Science\\
  Princeton University \\
  \texttt{\{allen.ren, bgovil, ty3, karthikn, ani.majumdar\}@princeton.edu} \\
}

\begin{document}
\maketitle

\setlength{\abovedisplayskip}{2pt}
\setlength{\belowdisplayskip}{2pt}
\setlength{\abovedisplayshortskip}{1pt}
\setlength{\belowdisplayshortskip}{1pt}

%===============================================================================

\begin{abstract}
Robust and generalized tool manipulation requires an understanding of the properties and affordances of different tools. We investigate whether linguistic information about a tool (\eg its geometry, common uses) can help control policies adapt faster to new tools for a given task. We obtain diverse descriptions of various tools in natural language and use pre-trained language models to generate their feature representations. We then perform language-conditioned  meta-learning to learn policies that can efficiently adapt to new tools given their corresponding text descriptions. Our results demonstrate that combining linguistic information and meta-learning significantly accelerates tool learning in several manipulation tasks including pushing, lifting, sweeping, and hammering. \footnote{Video showing the four manipulation tasks: \url{https://youtu.be/KTd6y-rf1Mk}}
\end{abstract}

\keywords{Language for Robotics, Tool Manipulation, Meta-learning} 

%===============================================================================
% \begin{bibunit}[corlabbrvnat]

\vspace{-5pt}
\begin{figure}[h]
\begin{center}
\includegraphics[width=1\textwidth]{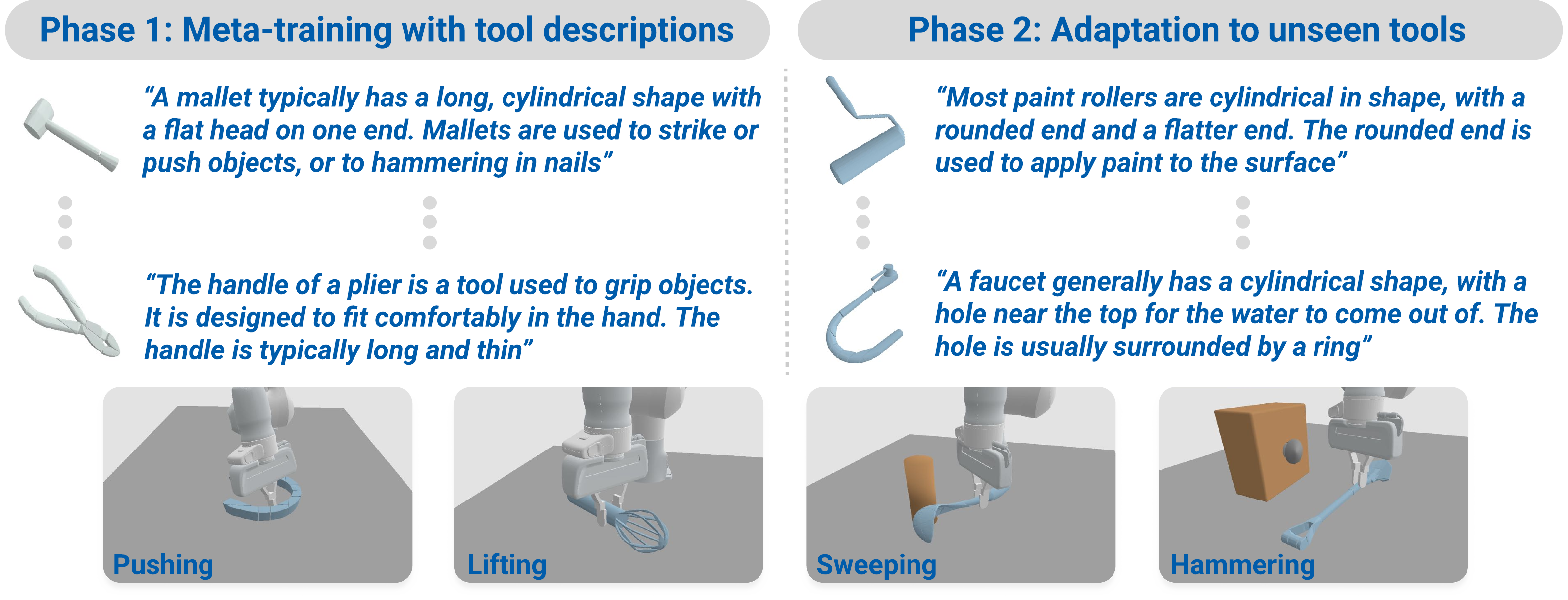}
\caption{Rich, semantic knowledge from language descriptions, such as geometric features and common use of the tools, can help policies adapt faster to unseen tools (right) in pushing, lifting, sweeping, and hammering tasks (bottom) after meta-learning on training tools (left).}
\label{fig:anchor}
\end{center}
\vspace{-18pt}
\end{figure}

\vspace{-5pt}
\section{Introduction}
\label{sec:introduction}
\vspace{-5pt}

The ability to quickly learn how to use a broad range of new tools is considered one of the hallmarks of human intelligence~\citep{taylor2014there}. Humans are able to attain a degree of proficiency with a new tool (\eg a new hand tool or a virtual tool such as a joystick or remote) within just a few minutes of interaction~\cite{gibson2002evolution}. This rapid learning relies on the ability to understand or discover the \emph{affordances}~\cite{gibson1977theory} of a new tool, \ie the ability to perform a certain action with the tool in a given environment. For example, a hammer affords the opportunity to grasp it, use it to hammer a nail, or use it to push another object. In contrast, a spatula affords the opportunity to flip a pancake or sweep food ingredients into a bowl. Endowing robots with the ability to quickly discover and exploit affordances of a new tool in order to learn how to perform a given task has been a long-standing grand challenge in robotics~\cite{Mason18}. 

While humans routinely rely on geometric priors and visual observations for tool manipulation, they can also `read' text manuals or linguistic descriptions to understand affordances of new tools and quickly adapt to using them. In this work, we investigate whether \emph{language} can help robots accelerate the process of learning to use a new tool. In particular, consider the following descriptions of two types of tools:

\begin{small}
% \vspace{-5pt}
\begin{displayquote}
%\quad\quad\quad``
\textit{The shape of tongs is typically that of a V, with two long, thin handles that come to a point at the top, and a gripping area in the middle.}

% \quad\quad\quad``
\textit{A spatula is a kitchen utensil for flipping over food while cooking. The head of the spatula is usually rectangular or oval in shape. The handle of the spatula is usually long and thin.}
\end{displayquote}
\end{small}

Our key intuition is that such natural language descriptions of tools contain information about the affordances of the tools, how to exploit these affordances for a given task, and how perceptual features of tools (\eg their visual appearance and geometry) relate to their affordances. Moreover, language can help capture the \emph{shared structure} of tools and their affordances. Thus, if one has previously learned to use a set of tools (with corresponding language descriptions), a description of a new tool can help to exploit this prior knowledge in order to accelerate learning. 

{\bf Statement of Contributions.} Our primary contribution is to propose \textbf{ATLA} --- \textbf{A}ccelerated Learning of \textbf{T}ool Manipulation with \textbf{LA}nguage --- a meta-learning framework leveraging large language models (LLMs) to accelerate learning of tool manipulation skills. The overall approach is illustrated in Fig.~\ref{fig:anchor}. 
% During meta-training time, we assume access to a set of tools and use a pre-trained LLM to automatically generate a diverse set of descriptions for each tool. In addition, we obtain the LLM feature representations corresponding to the language descriptions. 
ATLA utilizes LLMs in two distinct ways: to generate the language descriptions for tools and to obtain the corresponding feature representations. At meta-training time, the meta-learner updates a base-learner that quickly fine-tunes a manipulation policy; this fine-tuning process is conditioned on the LLM representations corresponding to the language descriptions of each tool. Specifically, we propose a simple gradient-based meta-learning setup based on Reptile \citep{nichol2018first} that performs off-policy updates. At test time, the base-learner adapts to a new tool using its language descriptions and interactions with it.
% ; this learning process is conditioned on the LLM feature representations of the descriptions of the tool. 
To our knowledge, our approach is the first to utilize LLMs to accelerate learning of new tools. We demonstrate the benefits of using language in a diverse set of tool-use tasks including pushing, lifting, sweeping, and hammering.  

%  and natural language descriptions for each tool

% Exploiting the affordances of a tool requires connecting perceptual features of the tool (\eg its visual appearance and geometry) to its function (in the context of a particular task). For example, using a hammer on a nail requires one to identify and grasp the handle, while pushing a hammer (without grasping it) requires approximately identifying its center of mass.

% Outline
% - Imagine robot that has learned how to use a set of tools... generalization and learning of new tools...
% - Tools can be similar in terms of affordances/kinematics/function
% - Intuition: language can provide hints of similarities...
% - Provide examples? 
% [dual use of language: generation of text descriptions and language representations...]
% - Contributions:
% - Pre-trained large language models can provide descriptions of tools and capture similarities...
% - Meta-learning setup: language-conditioned learning at the base step. 
% - Description of experiments and results

\vspace{-5pt}
\section{Related Work}
\label{sec:related}
\vspace{-2pt}

\parab{Tool Manipulation.}
Tool manipulation~\cite{kemp2007challenges, fang2020learning, xie2019improvisation, holladay2019force, qin2021rapidly} is one of the long-standing problems in robotics research.
A major challenge is understanding the affordances of the tool in different tasks.
%
% bohg2013data, bousmalis2018using, bicchi2000robotic, fearing1986implementing, stilman2007manipulation, levine2018learning yahya2017collective, fang2018multi, eisner2022flowbot3d
%
% This problem is important since a robot that intelligently uses tools to complete tasks is needed for applications such as personalized robotic assistants and unmanned underwater vehicles.
%
% Here, we present some recent works that are more relevant to ours.
%
% \cite{haarnoja2018composable} train a single Q-function for each manipulation task (pushing and merging objects) and then compose learned Q-functions for unseen tasks during evaluation.
%
% \cite{levine2018learning, nguyen2019review} use deep reinforcement learning (RL) to train a grasping policy while~\cite{melnik2019tactile} use other sensor information such as tactile sensors to improve learning efficiency.
%
% However, it is unclear whether the learned policy can be applied to new tasks or objects.
%
Previous work has modeled and learned affordances from parameterized keypoints on the tools \citep{fang2020learning, qin2021rapidly}, from human demonstrations \citep{xie2019improvisation}, and from spatial-temporal parsed graphs of the tools \citep{zhu2015understanding}.
% In addition, some prior works either focus on creating a 3D hand-object interaction dataset from demonstrated images or videos~\cite{cao2021reconstructing,hasson2021towards} or designing rendering engines~\cite{ehsani2021manipulathor}.
% %
% Furthermore, \cite{petrik2020learning} learns a manipulation policy by imitating the state trajectory estimated from demonstrated videos.
%
% The adoption of natural language for the generalization of tool manipulation is a key factor that distinguishes our work from the prior works.
Our work instead leverages natural language (\eg describing affordances of the tools in words) for generalization of affordances in tool manipulation and is compatible with previous approaches.

\parab{Language-informed Control.}
Natural language has been applied to enable efficient robotic learning through 
(1) generating primitive language instructions for producing control actions (\ie instruction following task)~\cite{yang2020robust, saycan2022arxiv, tellex2011understanding, matuszek2013learning, artzi2013weakly, chen2011learning, anderson2018vision, branavan2009reinforcement}, % sharma2021skill
(2) learning language-informed reward functions for training control policies~\cite{squire2015grounding, bahdanau2018learning, goyal2019using, williams2018learning, yang2021safe, sumers2021learning}, % fu2019language
and (3) using language to correct or adapt the behavior of the robot~\cite{co2018guiding, karamcheti2020learning}.
However, these works primarily translate natural language into action policies for a specific task with the text providing information on the desired actions that optimize returns (\eg ``\textit{push the door}'').
This means that the text is tightly coupled with the task seen during training, making it difficult to generalize to a new distribution of tasks with different dynamics.
In contrast, the text in our work only provides a high-level description of the property of each tool, encouraging the agent to extract useful information to generalize to a new task.
Some prior work~\cite{hanjie2021grounding, branavan2012learning, zhong2019rtfm, narasimhan2018grounding} also uses language descriptions of environment dynamics to enable generalization of policies but does not leverage meta-learning.

\parab{Meta-learning.}
Our work uses the framework of meta-learning~\cite{finn2017model,mishra2017simple,rajeswaran2019meta,ren2018meta,finn2018probabilistic}, in which the agent is trained with a distribution of tasks, and later adapts quickly to a previously unseen task.
``Reptile'' is proposed in \cite{nichol2018first} as a simple first-order, gradient-based meta-learning algorithm that learns an initialization of the neural network's parameters for fast adaptation at test time.
%
% \cite{santoro2016meta} then augment neural networks with memory to make accurate predictions after learning from a few samples. 
%
Recent papers~\cite{dubey2020connecting, vuorio2018toward, yoon2019tapnet, lan2019meta} have also explored providing additional context information (\eg the property of the task) to encode task-specific knowledge for a meta-learning agent.
However, all these works directly provide the context information either through scalar signals or a learned task embedding, which require domain expertise or a pre-training stage.
In this work, we assume that the agent is provided with a text description of the tool, which is more accessible and easier to collect.
\vspace{-5pt}
\section{Problem Formulation}
\label{sec:formulation}
\vspace{-5pt}

We consider the following goal: given a new tool and corresponding language description(s), we aim to \emph{quickly} learn a policy for using the tool to achieve a given task. We pose this problem in a meta-learning setting in which a policy is trained with a distribution of tools, and later adapts quickly to a previously unseen tool sampled from this distribution.

{\bf Meta-training.} During meta-training, we assume access to a set $\mathcal{T} = \{\tau_i\}_{i=1}^{K}$ of tools, where $K$ is the number of available tools. For each tool $\tau_i$, we are also provided a set of corresponding language descriptions $L_i = \{l_{ij}\}_{j=1}^{N_i}$, $l_{ij} \in \mathcal{L},$ where $N_i$ is the number of available descriptions for tool $\tau_i$ and $\mathcal{L}$ is the set of possible textual descriptions.
In addition, each $l_{ij}$ can describe a different property of tools such as shape and common use.
Given a particular robotic manipulator and a particular task (\eg pushing, lifting, sweeping, or hammering), each tool $\tau$ induces a partially-observable Markov decision process (POMDP): $\langle \mathcal{S}_\tau, \mathcal{A}, \mathcal{O}, \mathcal{P}_\tau, R_\tau \rangle$. Here, $\mathcal{S}_\tau$ is the state of the entire environment (\ie combined state of the robot, tool, and potentially other objects to be manipulated using the tool). The robot's action space $\mathcal{A}$ (\eg corresponding to robot joint torques) and observation space (\eg the space of RGB-D observations from a camera) are fixed across tools. The transition probabilities are given by $\mathcal{P}_\tau: \mathcal{S}_\tau \times \mathcal{A} \times \mathcal{S}_\tau \rightarrow [0,1]$, and the reward function is $R: \mathcal{S}_\tau \times \mathcal{A} \times \mathcal{S}_\tau \rightarrow [0,1]$. 
During meta-training, our goal is to learn a policy $\pi_\theta: \mathcal{O} \times \mathcal{L} \rightarrow \mathcal{A}$ parameterized by $\theta$ (\eg weights of a neural network) that can be quickly fine-tuned at test time. 

{\bf Meta-testing.} At test time, we are provided a new tool $\tau_\nu$ and corresponding language descriptions $L_\nu = \{l_{\nu j}\}_{j=1}^{N_\nu}, l_{\nu j} \in \mathcal{L}$. We aim to let the meta-learned policy quickly adapt to this new tool in a fixed number of interactions with the tool in order to maximize the expected cumulative reward. This is a challenging task since the new tool can be quite different in terms of visual appearance and affordances as compared to previously seen tools in meta-training.

% For each manipulation task $M_i$, we formulate a Markov Decision Process (MDP) $\langle \mathcal{S}, \mathcal{A}, \mathcal{P}, R \rangle$, where $\mathcal{S}$ denotes all the possible states, $\mathcal{A}$ denotes the action space, $\mathcal{P}: \mathcal{S} \times \mathcal{A} \times \mathcal{S} \rightarrow [0,1]$ denotes the transition probability, and $R: \mathcal{S} \times \mathcal{A} \times \mathcal{S}' \rightarrow [0,1]$ denotes the reward function. At training time, we assume access to a set of environments $E_i$, each consisting of a tool to be manipulated and a table-top robot manipulator. The environment is partially observable - we assume the robot has a sensor (\eg RGB camera) which provides observations $o_t \in \mathcal{O}$ of the environment ($o_t$ can include history observations). We denote additionally available information of the environment $l_E \in \mathcal{L}$ (\eg feature representation of the language description of the tool).

% For each task, we synthesize a policy $\pi_\theta: \mathcal{O} \times \mathcal{L} \rightarrow \mathcal{A}$ parameterized by $\theta$ (\eg weights of a neural network) that maximizes the expected cumulative reward as the training objective. After training, we aim to let the policy quickly adapt to unseen environments after a fixed number of interactions, which can be challenging if the test environment varies significantly from training ones (\eg distinct tool geometry). 

% \allen{introduce meta-learning objective here?}

\vspace{-5pt}
\section{Approach}
\label{sec:approach}
\vspace{-2pt}
The key idea behind our approach is to \underline{\smash{collect}} and \underline{\smash{embed}} language information of the environment into \underline{\smash{meta-learning}}, allowing the policy to adapt faster and better to unseen environments. 

\subsection{Collecting Language Information Using Pre-Trained Large Language Models}
\label{subsec:collect_lang}
\vspace{-2pt}
A common use of language in robotics is to use it to provide an instruction to the robot (\eg ``\textit{pick up the green block on the table}").
% Among applications of language use in robotics, the language often appears as the instruction (\eg ``pick up the green block on the table'') for the robot. 
Such instructions are typically specified by humans manually through crowd-sourcing, which can be labor intensive. 
In our setting, we consider language as additional information about the environment (\eg ``\textit{the hammer has a long handle and large head at the top}'').
The language here is not used to describe the goal (\eg what to do), but to provide information about properties of the environment (\eg tool shape).
This makes the text here \textit{task-agnostic}, forcing the agent to learn generalizable policies.
To obtain a diverse set of language descriptions, we are inspired by the recent advances in LLMs that are trained with vast amounts of online data and imbued with rich, semantic knowledge of different objects. We propose using LLMs to provide language descriptions of the tools in the form of \emph{question answering}. Specifically, we provide the GPT-3~\cite{brown2020language} model with the following template prompt through the OpenAI API:

\begin{footnotesize}
\begin{displayquote}
\quad\quad\quad``\textit{Describe the [feature] of [name] in a detailed and scientific response:} ''
\end{displayquote}
\end{footnotesize}

where ``feature'' is selected from one of [``shape'', ``geometry''] or one of  [``common use'', ``purpose''] and ``name'' describes the tool (\eg ``\textit{a hammer}'', ``\textit{a pair of tongs}''). We find that adding ``\textit{detailed and scientific}'' to the prompt significantly improves the quality of the texts generated. For each tool, we generate 10 different paragraphs of descriptions for each of the four features, and then combine paragraphs in each of the four permutations of the features (``shape'' and ``common use'', etc). Each tool $\tau_i$ is thus paired with a diverse set of 800 language descriptions $L_i$ (see Appendix \ref{app:tools} for more examples). Each description $l_{ij} \sim L_i$ is approximately 2-4 sentences long.

\vspace{-8pt}
\begin{figure}[h]
\begin{center}
\includegraphics[width=1\textwidth]{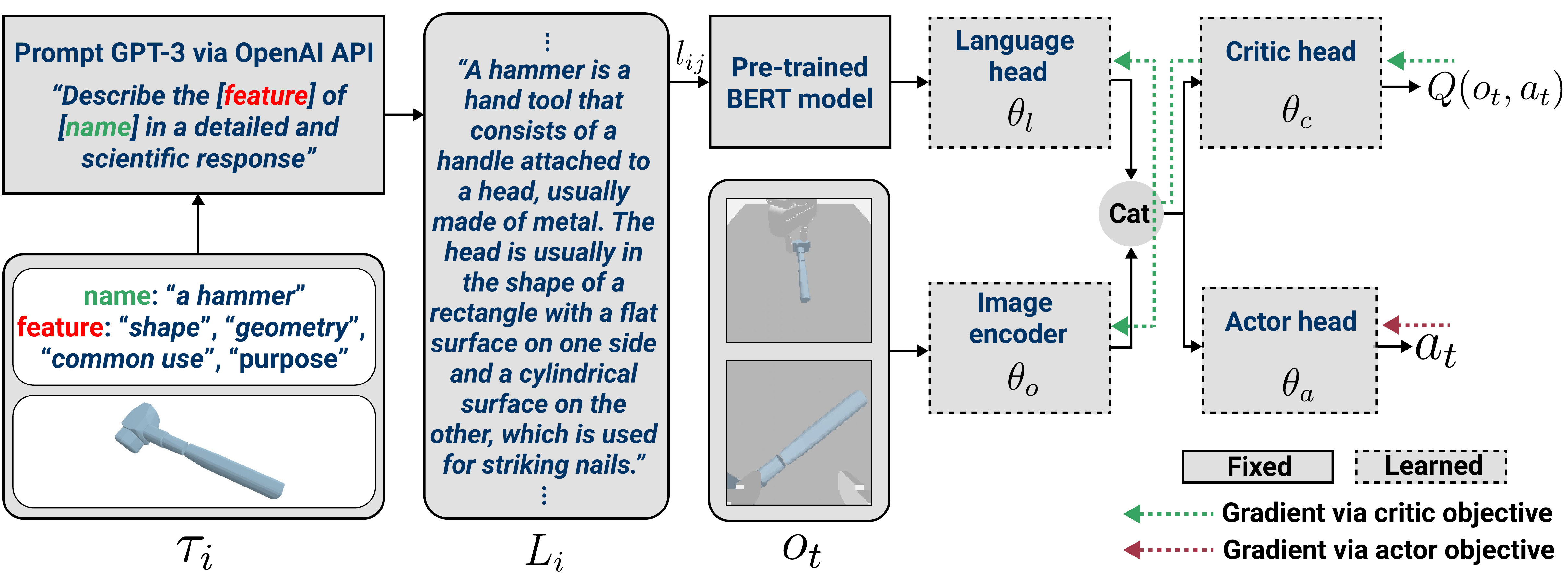}
\caption{\textbf{Model Overview.}
First, we prompt OpenAI GPT-3 to obtain a diverse set of language descriptions $L_i$ of the tool $\tau_i$. 
Then for each episode collected, we sample a language description $l_{ij}$ randomly from $L_i$, which is then fed into a pre-trained BERT model to obtain the representation.
%This allows the model to see a diverse set of text, which improves the generalization. 
%
The language head further distills the language information.
% (3) Next, for each episode, we sample a description $l_{ij}$ from $L_i$, and then feed it into a frozen BERT model into a single fully-connected layer $\theta_l.$ In addition, the image observations from two camera angles are fed into an image encoder $\theta_o.$
%
% (4) Finally,
We concatenate the representations from the language head and the image encoder, and then the features are shared by the critic head and the actor head.
% $\theta_l$ and $\theta_o$ and feed this representation into a critic head $\theta_c$ for producing a Q-function $Q(o_t,a_t)$ and a actor head $\theta_a$ for producing an action $a_t$.
%
%During training, the actor head is updated via the gradient of actor objective; the critic head, language head, and image encoder are updated by the gradient of critic objective.
}
\label{fig:model}
\end{center}
\vspace{-18pt}
\end{figure}

\vspace{-2pt}
\subsection{Obtaining Feature Representations from Large Language Models}
\vspace{-2pt}

\begin{wrapfigure}[14]{r}{0.40\textwidth}
\vspace{-20pt}
\begin{center}
    \begin{tabular}{@{}c@{}}
    \includegraphics[width=0.40\textwidth]{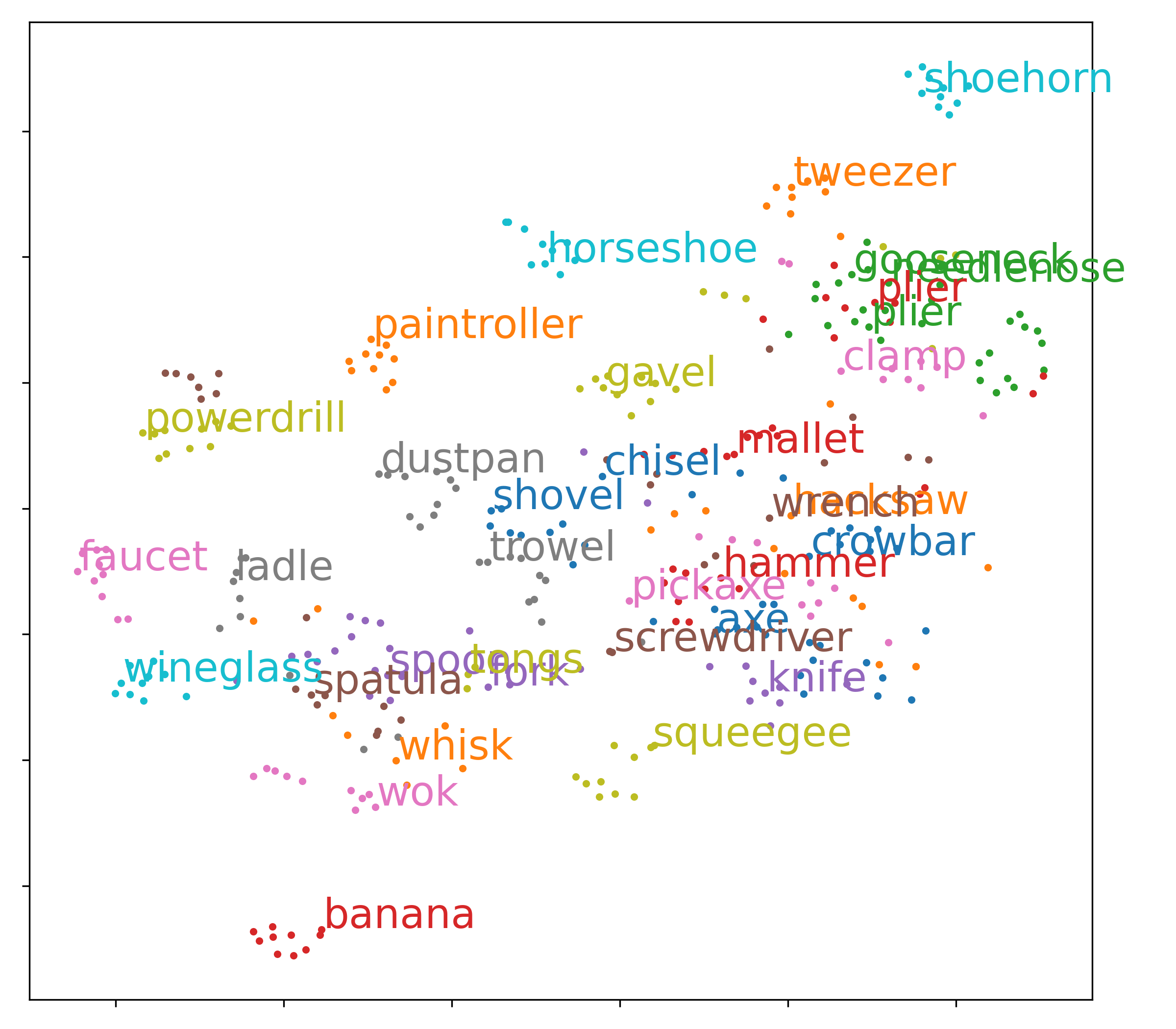} \\
    \end{tabular}
\end{center}
\vspace{-15pt}
\caption{T-SNE results of BERT output for multiple language descriptions of each tool.}
\vspace{+50pt}
\label{fig:tsne}
\end{wrapfigure}

With the collected language descriptions, we now incorporate them into policy training. One common choice is to train a language module (\eg long short-term memory (LSTM)~\cite{hochreiter1997long}) from scratch to embed features of the language input, which can take substantial time and effort to tune. Instead, we use a pre-trained LLM to distill the language descriptions into feature representations. Since LLMs are trained with vast amounts of data, they can better interpret and generalize to the diverse set of long descriptions. Since the GPT-3 model is not publicly available, we opt for the Google BERT-Base \citep{turc2019well} model on HuggingFace, which has 110.1M parameters and outputs a 768-dimensional vector representation for each description input. T-SNE analysis shown in Fig.~\ref{fig:tsne} demonstrates that, without any fine-tuning, the model already captures semantic meanings of the descriptions among tools (\eg hammer and mallet are close to each other).

Fig.~\ref{fig:model} shows the overall neural network architecture. 
We first prompt GPT-3 to obtain a set $L_i$ of text descriptions for tool $\tau_i$ via the procedure in Sec.~\ref{subsec:collect_lang}.
During meta-training, we randomly sample $l_{ij}$ from $L_i$ for each episode to ensure that the policy sees a diverse set of descriptions.
We then freeze the BERT model during policy training. 
The output from BERT is fed into a single fully-connected layer with ReLU (language head, $\theta_l$).
The image observations (possibly from two camera angles--one from overhead and one from the wrist) are passed through convolutional layers (image encoder, $\theta_o$), whose output is then concatenated with that from the language head. The actor head ($\theta_a$) and critic head ($\theta_c$) then output the action $a_t$ and the corresponding value for the Q function $Q(o_t, a_t)$. See Appendix \ref{app:model} for more details of the neural network setup.
%
%Next, we discuss the meta-training setup.
\vspace{-3pt}
\subsection{Meta-training and Testing Language-Conditioned Manipulation Policies}
% \parab{Meta-training}
\vspace{-2pt}

We hypothesize that additional language information of the tools promotes generalization. However, zero-shot transfer to unseen tools can be difficult given the distinct geometries and affordances. Thus we perform meta-training for explicitly training the policy to adapt to distinct tools. Our algorithm (shown in Algorithm 1) is based on Reptile \citep{nichol2018first}, a simple first-order gradient-based meta-learning algorithm, but with an additional sampling strategy to prevent overfitting.
At each iteration of the meta-training, one tool $\tau$ is sampled from the training set $\mathcal{T}.$
At the base level (line 5), we run the current policy with the language description $l_{ij}$ sampled from $L_i$.
We then add the collected experiences in a base replay buffer $\beta_\text{base},$ 
and perform $B$ iterations of off-policy updates using Soft Actor Critic (SAC) \citep{haarnoja2018soft} in order to obtain the final policy parameters $\vtheta'=[\theta_l', \theta_o', \theta_a', \theta_c'].$
%
% Note that the critic objective is used to update $\theta_o, \theta_l,$ and $\theta_c$ while the actor objective is used to update $\theta_a.$ 
%
%
Then at the meta level (line 12), the network is updated with a gradient step towards $\theta'$:
% \vspace{-5pt}
\begin{equation}
    \vtheta_\text{new} \leftarrow \vtheta + \alpha (\vtheta' - \vtheta),
\label{eq:meta}
\end{equation}
where $\alpha$ is the meta-learning rate and $\vtheta$ is the collection of the old policy parameters $\theta_l, \theta_o, \theta_a, \theta_c.$ 
We also highlight the following remarks including differences to the Reptile algorithm: 
\begin{itemize}[leftmargin=*]
\setlength\itemsep{0em}
    \item To reduce variance, the actor and critic share the parameters of language head $\theta_l$ and the image encoder $\theta_o$, and both modules are updated only with the critic objective.
    \item To prevent overfitting to language descriptions of one tool during training, experiences collected from all tools are saved in a meta replay buffer $\beta_\text{meta}$ (line 14), and $\beta_\text{base}$ for each tool is initialized with random samples from $\beta_\text{meta}$ (line 3). During policy update at base training, $30\%$ of the experiences are sampled from $\beta_\text{meta}$. During test time, we do not use any experiences from $\beta_\text{meta}$. We demonstrate the effectiveness in Sec.~\ref{sec:results}.
    \item In applications of Reptile in supervised learning, the meta update is often averaged over $N$ sampled tasks: $\theta_\text{new} \leftarrow \theta + \alpha \sum_i^N (\theta'_i - \theta)$. However, we find that $N=1$ trains faster and also matches the test objective of adapting to a single tool.
    \item Performing only a single meta update after $B$ iterations of base update can be inefficient. Instead of using a large learning rate $\alpha$ which causes unstable training, we perform $M=2$ meta updates for each adaptation to one tool, but collect experiences to $\beta_\text{base}$ only at the first update. 
    % In Sec.~\ref{sec:results} we show that this empirically achieves better training efficiency.
    % See Fig.~\ref{fig:?} for visualization. \allen{add a simple figure for gradient updates}
\end{itemize}
% \parab{Test time.} 
Algorithm 2 shows the procedure of adaption at test time.
First, the model is provided with a target test tool $\tau_\nu$ and a set of language descriptions.
With the adaptation budget $B_\nu,$ we run the policy with the language description $l_j$ sampled from $L_\nu.$
The collected experiences are stored in the buffer $\beta_\text{base}$ and used to update the policy parameters.

\vspace{-5pt}
\begin{algorithm}[H]
\caption{ATLA: Meta-training, $N=1$}
\begin{algorithmic}[1]
    \Require $\mathcal{T} = \{\tau_i\}_{i=1}^{K}$: training set of tools; $\{L_i\}_{i=1}^K$: sets of language descriptions; $\theta_l, \theta_o, \theta_a, \theta_c$: policy modules; $\beta_\text{meta}$: meta replay buffer.
    \While{meta-training}
        \State Sample $\tau$ from $T$ \quad\texttt{\# meta level}
        \State Reset $\beta_\text{base}$ with samples from $\beta_\text{meta}$ % \Comment{Skip at test time}
        \For{$m = 1 \ \text{to} \ M$}
            \For{$b = 1 \ \text{to} \ B$} \quad\texttt{\# base level}
                \If{$m = 1$}
                    \State Collect episodes each with $l_{ij} \sim L_i$; add to $\beta_\text{base}$
                \EndIf
                \State Sample from $\beta_\text{base}$ and update $\theta_o, \theta_l, \theta_c$ with the critic objective
                \State  Sample from $\beta_\text{base}$ and update $\theta_a$ with the actor objective
            \EndFor
        \State Meta update $\theta_o, \theta_l, \theta_c, \theta_a$ with Eq.~\ref{eq:meta}
        \EndFor
        \State Add $\beta_\text{base}$ to $\beta_\text{meta}$
    \EndWhile
	\end{algorithmic}
\normalsize
\end{algorithm}
\vspace{-5pt}

\vspace{-10pt}
\begin{algorithm}[H]
\caption{Adaption at test time}
\begin{algorithmic}[1]
    \Require $\tau_\nu$: test tool; $L_\nu$: set of language descriptions; $\theta_l, \theta_o, \theta_a, \theta_c$: policy modules; $\beta_\text{base} \leftarrow \emptyset$: base replay buffer
    \For{$b = 1 \ \text{to} \ B_\nu$}
        \State Collect episodes each with $l_j \sim L_\nu$; add to $\beta_\text{base}$
        \State Sample from $\beta_\text{base}$ and update $\theta_o, \theta_l, \theta_c$ with the critic objective
        \State Sample from $\beta_\text{base}$ and update $\theta_a$ with the actor objective
    \EndFor
	\end{algorithmic}
\normalsize
\end{algorithm}

\section{Experiment Setup}
\label{sec:experiment-setup}
\vspace{-5pt}

Through different tool manipulation tasks in simulation, we aim to investigate the following questions: (1) Does language information help achieve better adaptation to new tools? (2) Does meta-learning improve adaptation to new tools? (3) How does the choice of pre-trained LLMs affect policy training? (4) Does language information help the policy utilize tools' affordances effectively?

\parab{Tasks.} Four different tool manipulation tasks are implemented (see bottom of Fig.~\ref{fig:anchor}): (1) pushing: pushing the tool to a fixed location on the table; (2) lifting: reaching and lifting the tool up from the table to some target height; (3) sweeping: using the tool to sweep a cylinder to a fixed location on the table; (4) hammering: using the tool to hammer a peg further into a hole in a block. Solving these tasks benefits from an understanding of the geometric affordances of the tools such as the grasp location. See App.~\ref{app:tasks} for more details of the task setup including the reward functions.

\parab{Robot.} We build custom simulation environments with a 7-DOF Franka Panda arm in the PyBullet simulator \cite{coumans2016pybullet}. We use RGB cameras with $128\times128$ image outputs, placed at different off-arm locations and at the arm wrist depending on the needs of the tasks. For all tasks, we use 4-DOF cartesian velocity (3D translation and yaw) as the action output from the policy. The arm joints are then commanded with a jacobian-based velocity controller at 5Hz. The policy does not command the gripper; instead, we use the heuristic that once the gripper is below some height, the gripper closes to grasp the object. If the grasp fails, the gripper re-opens if it rises above the threshold.

\parab{Tools.} We collect a total of 36 objects (See App.~\ref{app:tools} for the full list) from open sources. Most of the objects are common tools such as a hammer and an axe. Some of them are less used as tools but have distinct geometry and affordances, such as a banana whose inner curvature may help push other objects. We split the objects into a training set of 27 and a test set of 9 --- we try to separate objects with similar geometry or affordances (\eg hammer and mallet) into different sets.

\parab{Baselines.} For all tasks we compare ATLA (ours) with the following baselines (Table~\ref{fig:results}): \textbf{(a)} \emph{AT-TinyLA} (ours): ATLA with a smaller BERT encoder (BERT-Tiny \cite{turc2019well} with 4.4 million parameters and 128-dimensional output). 
% We include this baseline to test whether scaling laws happen in our task. 
\textbf{(b)} \emph{AT}: ATLA without language information. 
% This is to show the importance of the language information for meta-learning. 
\textbf{(c)} \emph{AT-XL}: ATLA without language information but larger networks for $\theta_a$ and $\theta_c$ (matching the number of parameters of ATLA).
%This is to show that although we have the same representation capability of the network, the network that is capable of processing linguistic information can have better generalization.
\textbf{(d)} \emph{SAC-LA}: vanilla multi-environment training with SAC and language information but without meta-training objective. 
% This is to show that mete-training could enable better adaptation.
\textbf{(e)} \emph{SAC}: SAC-LA but without language information.

\parab{Metric.} For all experiments, we save the model checkpoint with the highest running-average reward on the training dataset. After training, for each test tool we load the checkpoint and run a fixed number of iterations of adaptation. In Fig.~\ref{fig:results}, we report the highest reward at adaptation, averaged over 3 seeds for each test tool. See Appendix \ref{app:results} for reward in numbers for each tool and task.

\vspace{-5pt}
\section{Results}
\label{sec:results}

% Fig.~\ref{fig:curves} compares the adaptation curves between ATLA and AT: ATLA usually has a higher base reward, and also adapts faster to higher reward.

\begin{figure}[h]
\begin{center}
\includegraphics[width=1.0\textwidth]{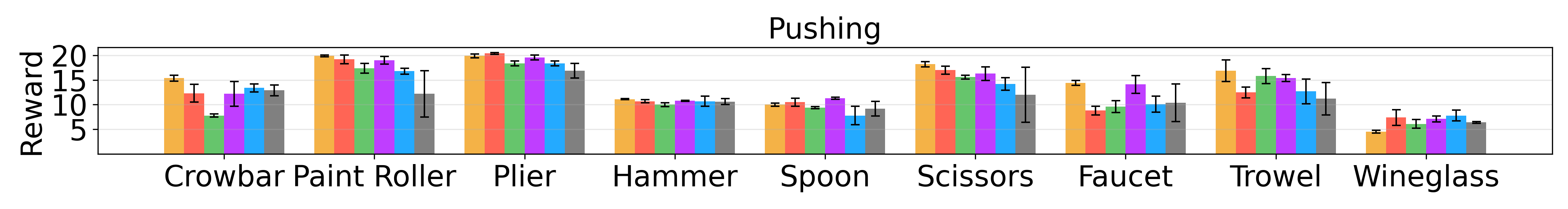}
\vspace{-1mm}
\includegraphics[width=1.0\textwidth]{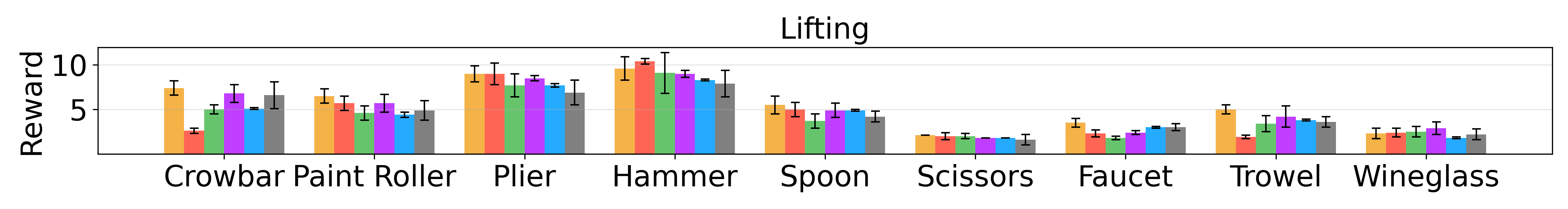}
\vspace{-1mm}
\includegraphics[width=1.0\textwidth]{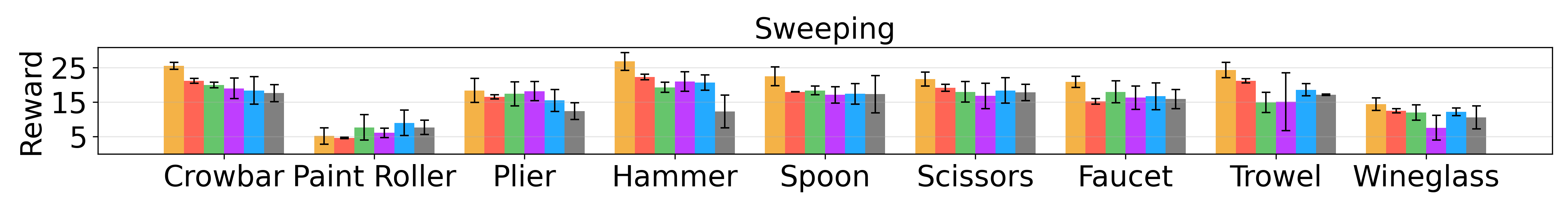}
\vspace{-1mm}
\includegraphics[width=1.0\textwidth]{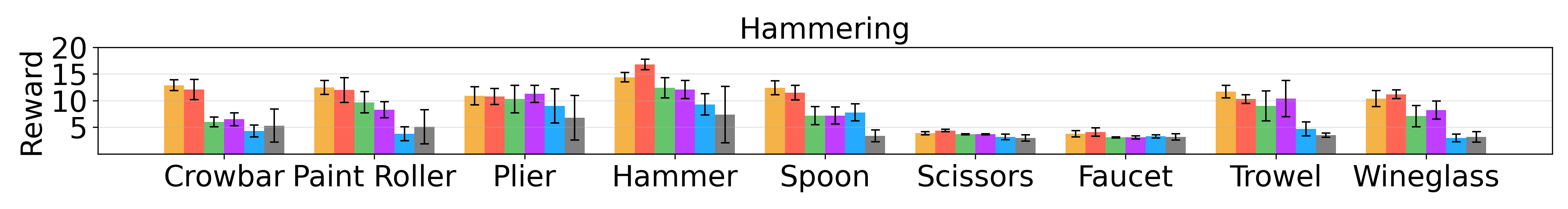}
\vspace{-1mm}
\includegraphics[width=0.65\textwidth]{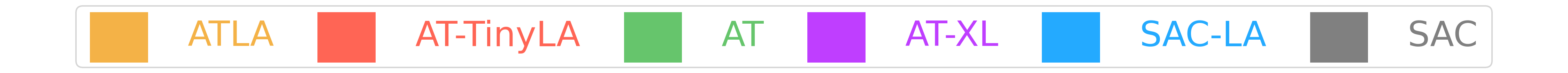}
\caption{Post-adaptation reward in mean and standard deviation over 3 seeds across 4 tasks and 9 test tools.}
\label{fig:results}
\end{center}
\vspace{-15pt}
\end{figure}

\vspace{-10pt}
\begin{figure}[h]
\begin{center}
\includegraphics[width=1\textwidth]{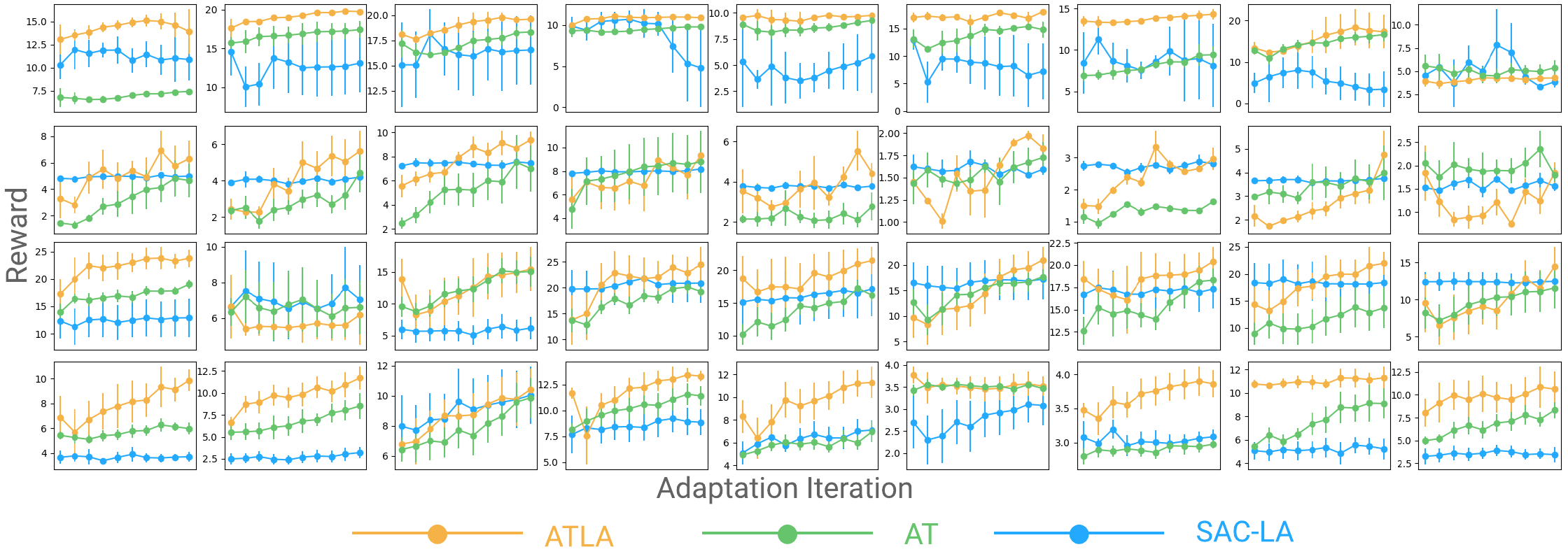}
\caption{Curves of adaptation iteration vs. reward of the test tools (columns) in the four tasks (rows).}
\label{fig:curves}
\end{center}
\vspace{-28pt}
\end{figure}

\parab{Q1: Does language information help achieve better adaptation to new tools?} Fig.~\ref{fig:results} shows that ATLA and AT-TinyLA perform better than AT and AT-XL among most tools in the 4 tasks. The differences are more significant in sweeping and hammering, which are more difficult and language information can better inform the affordances of the tools. Among the 9 tools, ATLA and AT-TinyLA always perform better with crowbar, plier, hammer, scissors, faucet, and trowel. ATLA and AT-TinyLA do not perform better mostly when low reward is achieved for all baselines for that tool, such as wineglass in pushing and lifting and paint roller in sweeping. Fig.~\ref{fig:curves} also shows that language helps faster learning in ATLA compared to AT in most cases, with the agent achieving higher rewards with fewer episodes of adaptation. 

We also find that different tools learn better for different tasks. For example, hammer is better in sweeping than plier probably due to its long bar, but worse in pushing also due to the small inertia along the long bar causing instability during pushing. As ATLA performs better among most tools, language information can provide useful information about the tool affordances in different tasks. 
% \parab{Ablation: sampling language description from the set.} \TODO{finish}

\parab{Q2: Does meta-learning improve adaptation to new tools?} Table~\ref{fig:results} shows that with or without language information (ATLA / AT-TinyLA vs. SAC-LA, or AT / AT-XL vs. SAC), meta-learning improves final performance after adaptation. Without meta-learning, SAC-LA shows smaller improvement over SAC (\eg plier, hammer, scissors, and trowel). This demonstrates that language information particularly helps when combined with meta-learning.
\begin{wrapfigure}[9]{r}{0.32\textwidth}
\vspace{-13pt}
\begin{center}
    \begin{tabular}{@{}c@{}}
    \includegraphics[width=0.32\textwidth]{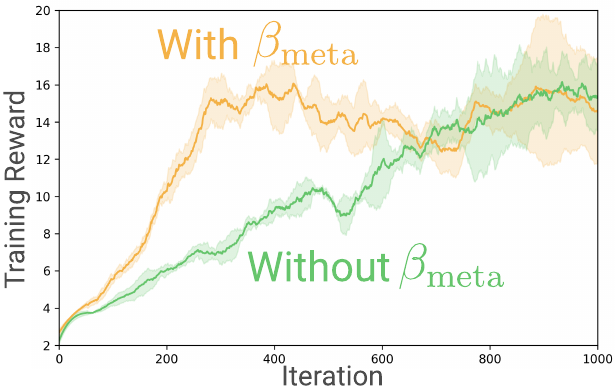}
    \end{tabular}
\end{center}
\vspace{-15pt}
\caption{Using meta replay buffer also accelerates training.}
\vspace{+0pt}
\label{fig:ablation-reward}
\end{wrapfigure}
Fig.~\ref{fig:curves} also compares the adaptation curves between ATLA and SAC-LA: those of SAC-LA tend to stagnate or fluctuate while those of ATLA tend to rise steadily. This indicates meta-learning trains the policy to better adapt to new tools after training.

\parab{Ablation: meta replay buffer.} We investigate the effect of re-using experiences from other tools (saved in $\beta_\text{meta}$ during adaptation). For this, we run ATLA without $\beta_\text{meta}$ on the pushing task. Across the 9 test tools, the post-adaptation reward is mostly lower ($-20\%, -10\%, -13\%, -18\%, +5\%, -25\%, -18\%, -18\%, -5\%$) compared to ATLA with $\beta_\text{meta}$. Note that the effect is more prominent when the reward difference between ATLA and AT is larger (\eg 20\% with crowbar and 25\% with scissors), indicating that language information is more effective if $\beta_\text{meta}$ is applied. We also find using $\beta_\text{meta}$ accelerates the meta-learning process (Fig.~\ref{fig:ablation-reward}) --- demonstrating that sharing experiences among tools makes training more efficient. 

\parab{Q3: How does the choice of pre-trained LLMs affect policy training?} Table \ref{fig:results} shows that ATLA usually attains higher post-adaptation reward than AT-TinyLA, which uses a smaller pre-trained BERT model, indicating that the policies benefit from the richer representation of the language descriptions that the bigger BERT model offers.

\begin{figure}[h]
\centering
\captionsetup{width=0.6\textwidth}
\begin{minipage}{.70\textwidth}
  \centering
  \includegraphics[width=0.95\linewidth]{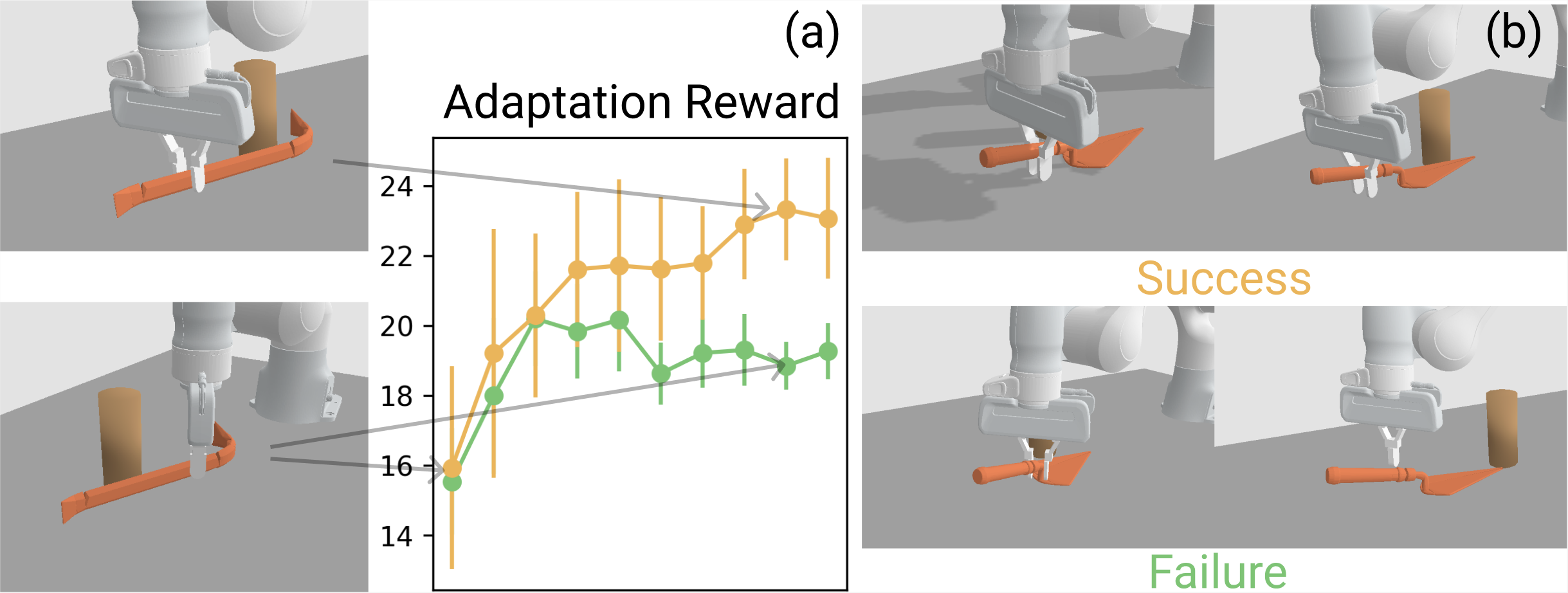}
  \captionof{figure}{With language information, ATLA is able to adapt the policy to utilize the affordances of the tools -- (a) curved hook on a crowbar; (b) handle on a trowel.
  %curved hook of the crowbar for better control of the object being swept.
  }
  \label{fig:crowbar}
\end{minipage}%
\begin{minipage}{.30\textwidth}
  \centering
  \captionsetup{width=1\linewidth}
  \includegraphics[width=0.70\linewidth]{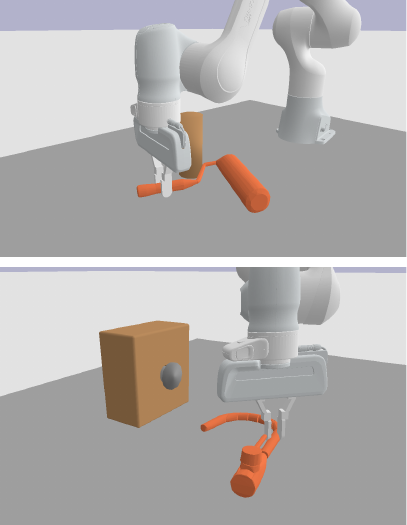}
  \captionof{figure}{Policies may fail to correctly utilize affordances.}
  \label{fig:failure}
\end{minipage}
\vspace{-1mm}
\end{figure}

\parab{Q4: Does language information help the policy utilize tools' affordances effectively?} The results above have shown that language descriptions of the geometric features and common use of the tools help policies adapt to new tools for a given task. Fig.~\ref{fig:crowbar}(a) visualizes the effect in the example of using a crowbar for sweeping. Language descriptions of a crowbar often contain phrases including ``long and thin bar'', ``curved'', ``hook'', ``used to leverage'', and ``used to pry open things''. With the descriptions, ATLA (orange curve in Fig.~\ref{fig:crowbar}(a)) enables the policy to adapt quickly to this tool unseen during meta-training --- the policy learns to use the curved hook to better steer the cylinder towards the target. As a comparison, we replace the descriptions with only the sentence ``A crowbar is a long and thin bar,'' and the policy (green curve in Fig.~\ref{fig:crowbar}(a)) does not adapt as well.

One common feature among tools is the handle. Language descriptions of a trowel includes phrases like ``flat, triangular blade'', ``handle to be grasped'', and ``used for scooping''. While ATLA learns to grasp at the handle (Fig.~\ref{fig:crowbar}(b) top), when we remove ``handle'' from all the descriptions, the robot fails to grip firmly on the handle and loses the grip eventually (Fig.~\ref{fig:crowbar}(b) bottom).

% \vspace{-2pt}

\parab{Limitations.} In some cases, we observe that the policies can still fail to correctly utilize the tool affordances with language information. In the sweeping task, the paint roller is the only tool that ATLA fails to perform the best with. Fig.~\ref{fig:failure} (top) shows the grasp learned by the policy. It failed to use the bigger opening between the roller and the handle on the other side to sweep the cylinder. Fig.~\ref{fig:failure} (bottom) shows that in the hammering task, the policy fails to use the faucet, specifically, its relatively flat head to push towards the nail. However, these affordances can be sensitive to the initial pose of the tool and can be difficult to explore. This also highlights one of the limitations of our work. We use a relatively simple task policy setup for tool manipulation tasks, which is directly mapping image inputs to Cartesian velocity commands. This creates challenges in exploration and learning the skills even with language information. One remedy is to combine with approaches like keypoint-based methods \citep{fang2020learning} that inject additional domain knowledge into the policy.
In addition, we use a relatively small dataset of tools, which may limit the potential of using language information. It would be particularly interesting to model revolute joints of the tools and perform more complex tasks such as picking up objects with a pair of tongs.
Furthermore, our current evaluation does not consider real robot experiments, which is \textit{not} the focus of this work. However, we expect that our approach can generalize to the real world without significant changes since language information is not involved in simulation and the arm does not operate near its dynamics limits.

% there is no agile or fast movement of the arm involved, making the sim-to-real gap small.

% A slightly surprising finding is that meta-learning also helps training performance. In the hammering task, SAC-LA and SAC achieves very low reward
% Fig.~\ref{fig:?} shows the curves for reward for training tools during training \TODO{add training curves for training curves too} -- without meta-learning, often vanilla multi-environment SAC training does not train as well.
% \parab{Ablation: number of meta update per adaption to tool.} We run experiments using $M=1$, performing one meta update after each base-level adaptation instead of two, in pushing task. Fig.~\ref{fig:ablation-meta-update} compares the reward of training tools during meta-training. With $M=2$, the policy learns significantly faster.

\vspace{-5pt}
\section{Conclusion}
\label{sec:conclusion}
\vspace{-2pt}
In this work, we investigate using large language models (LLMs) to accelerate adaptation of policies to new tools in tool manipulation tasks.
We use LLMs to both (1) generate diverse language descriptions of the tool geometry and common use, and (2) obtain vector representations of the descriptions. We then propose language-conditioned meta-learning that trains policies to quickly adapt to new tools. The results demonstrate that combining language information and meta-learning significantly improves the performance when adapting to unseen tools. 

%
%\parab{Limitation.}
%(1) We use a simple end-to-end setup for the task policy, as elaborated in Sec.~\ref{sec:experiment-setup}.
%
%(2) We use a relatively small dataset of tools, which may limit the potential of using language information. It would be particularly interesting to model revolute joints of the tools and perform more complex tasks such as picking up objects with a pair of tongs.
%
% (3) We do not consider real robot experiments, which is not the focus of this work.

% In addition, we would also like to extend our work to more complex tasks as such as moving prismatic (\eg opening a drawer) and revolute joints (\eg closing a door).
%
% This requires predicting how the object moves depending on the force applied.
% %
% First, our current evaluation does not consider real robot experiments, which is not the focus of this work. However, we expect that our algorithm can generalize to the real world since our simulator is accurate and there is no agile or fast movement of the arm involved, making the sim-to-real gap small.

% Furthermore, we observe some failure modes of the algorithm such as XXX. \jimmy{add failure modes here}
%
% This is can be mitigated by XXX.
%
% We leave this as future work.

%===============================================================================

\clearpage
% The acknowledgments are automatically included only in the final and preprint versions of the paper.
\acknowledgments{The authors were partially supported by the Toyota Research Institute (TRI), the NSF CAREER Award [\texttt{\#}2044149], the Office of Naval Research [N00014-21-1-2803, N00014-18-1-2873], and the School of Engineering and Applied Science at Princeton University through the generosity of William Addy ’82. This article solely reflects the opinions and conclusions of its authors and not NSF, ONR, Princeton SEAS, TRI or any other Toyota entity.}

%===============================================================================

\bibliography{bib-lang}  % .bib
% \putbib[bib-lang]  % .bib
% \end{bibunit}

%==================================================
\renewcommand{\thetable}{A\arabic{table}}
\renewcommand{\theequation}{A\arabic{equation}}
\renewcommand\thefigure{A\arabic{figure}}
\renewcommand{\thesubsection}{A\arabic{subsection}}
\setcounter{figure}{0}
\setcounter{table}{0}
\setcounter{equation}{0}

\clearpage
% \begin{bibunit}[corlabbrvnat]
\section*{Appendix}
\label{sec:appendix}

\subsection{Tool Information}
\label{app:tools}
Table \ref{app:table:tools} shows additional information about the tools used in the paper, including a sample view of the object mesh, sample language descriptions, and the train-test split.

% \scalebox{0.85}{
\begin{footnotesize}
% \begin{longtable}{ c{0.10\textwidth} c{0.10\textwidth} c{0.50\textwidth} c{0.10\textwidth} }
\begin{longtable}{cccc}
% \begin{table*}[h]
% \centering
% \vspace{0.0in}
% \scalebox{0.70}{
% \begin{tabular}{cccc}
\toprule
% \hline \\
Tool & Sample view & Sample Language description & Used for testing \\  \hline
    \multirow{1}{*}{axe} & \Includegraphics[height=10mm]{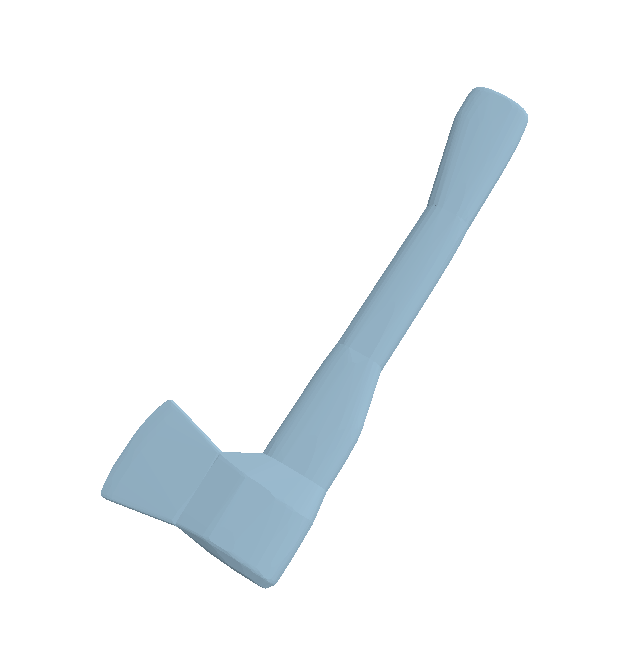} & \begin{minipage}[t]{0.5\columnwidth} \vspace{-4mm} An axe typically has a long, cylindrical handle with a flared end. The head of the axe typically has a slightly curved blade. An axe is often used for chopping wood. \end{minipage} & No \\
    \multirow{1}{*}{chisel} & \Includegraphics[height=8mm]{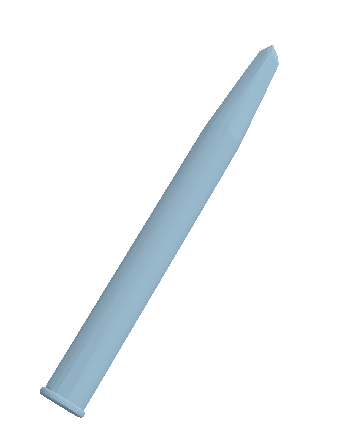} & \begin{minipage}[t]{0.5\columnwidth} \vspace{-4mm} Chisel can make clean, precise cuts using the beleveled edge. A chisel is a hand tool with a blade attached to a handle.  \end{minipage} & No \\
    \multirow{1}{*}{crowbar} & \Includegraphics[height=7mm]{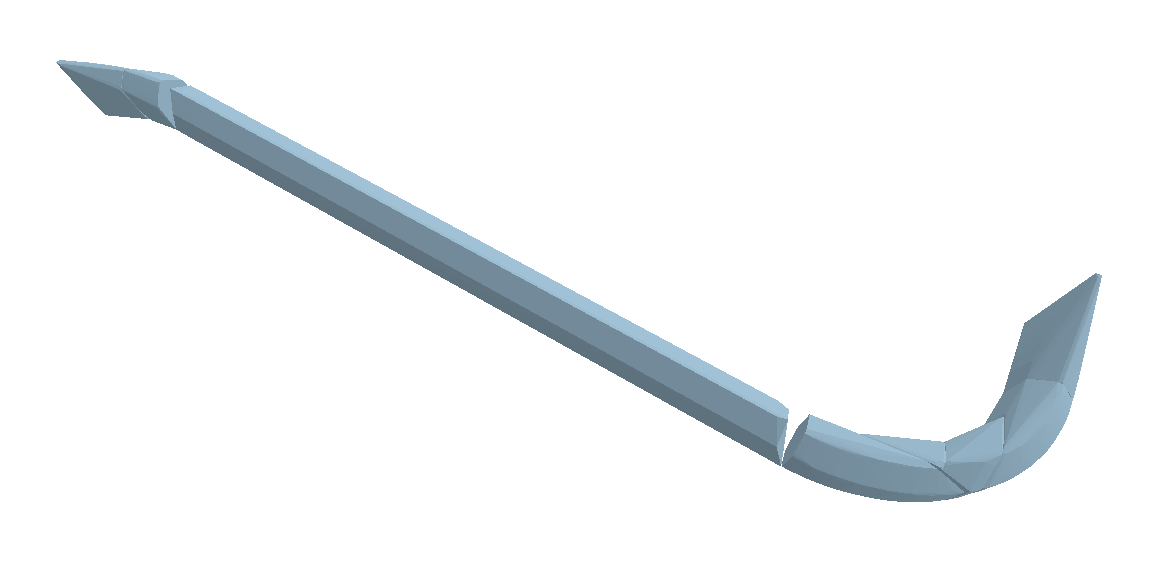} & \begin{minipage}[t]{0.5\columnwidth} \vspace{-4mm} A crowbar is used as a lever to pry things open. One end of a crowbar is usually curved or hooked so that it can be jammed under an object to apply leverage. \end{minipage} & Yes \\
    \multirow{1}{*}{ shovel} & \Includegraphics[height=9mm]{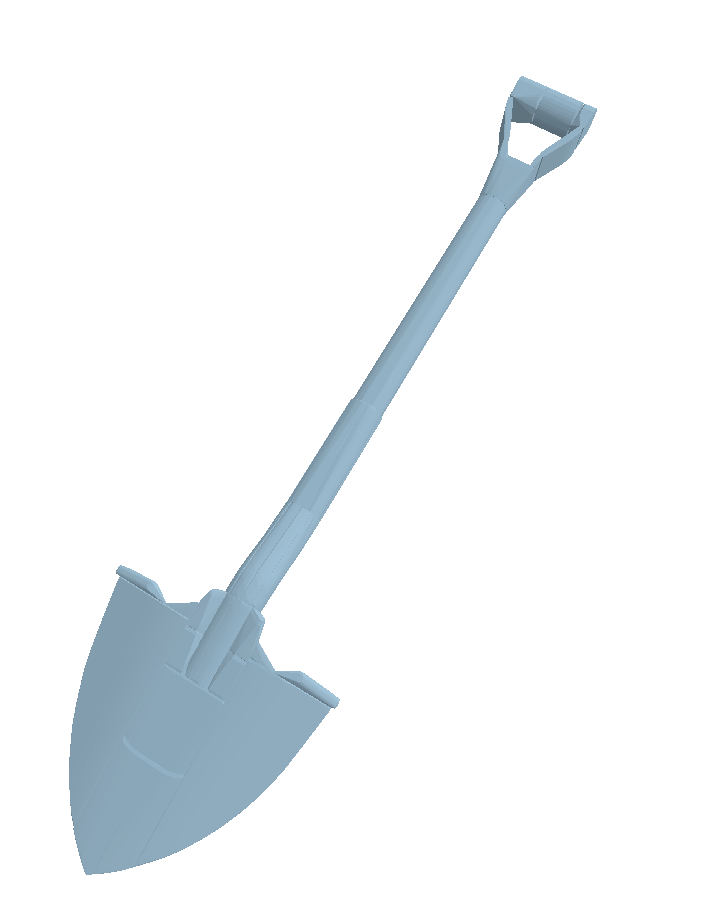} & \begin{minipage}[t]{0.5\columnwidth} \vspace{-4mm} A shovel has a long, cylindrical handle and a scoop-shaped blade. The shovel typically has a curved part for digging into and scooping up materials. \end{minipage} & No \\
    \multirow{1}{*}{ hacksaw} & \Includegraphics[height=10mm]{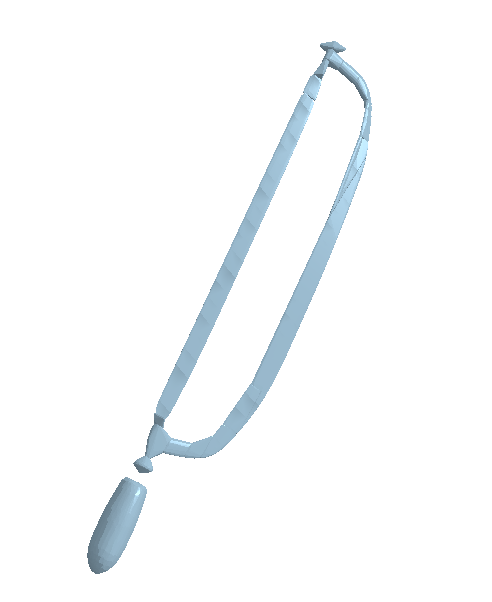} & \begin{minipage}[t]{0.5\columnwidth} \vspace{-4mm} A hacksaw is a hand saw with a thin blade attached to a handle, used for cutting various such as metal, plastic, or wood. A hacksaw is a saw with a thin, toothed blade on a rigid frame, used for cutting wood or metal. \end{minipage} & No \\
    \multirow{1}{*}{ paintroller} & \Includegraphics[height=9mm]{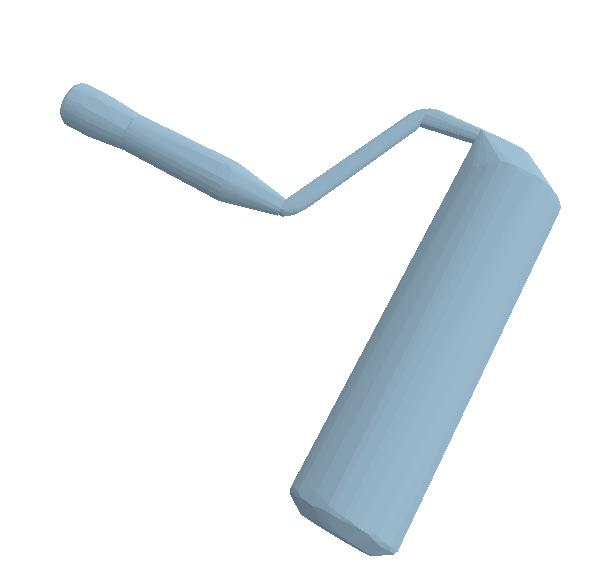} & \begin{minipage}[t]{0.5\columnwidth} \vspace{-4mm} One common use of a paint roller is to apply paint evenly to a surface such as walls or ceilings. A paint roller consists of a long, cylindrical body with a handle on one end. \end{minipage} & Yes \\
    \multirow{1}{*}{ tweezer} & \Includegraphics[height=10mm]{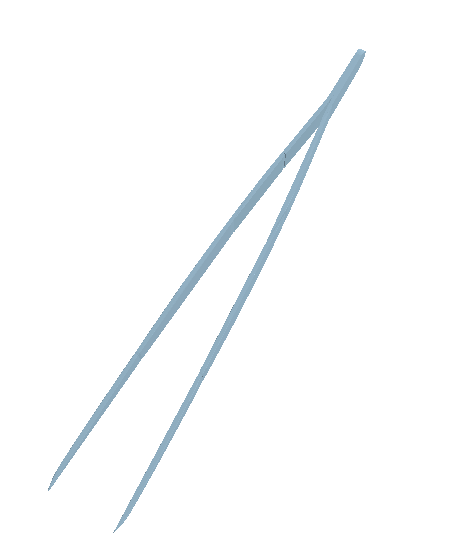} & \begin{minipage}[t]{0.5\columnwidth} \vspace{-4mm} A tweezer is a hand-held tool with two arms that meet at a point. A tweezer is a small hand-held tool with two pointed jaws that are used to pick up small objects or to remove unwanted hair or debris from the body. \end{minipage}  & No \\
    \multirow{1}{*}{ whisk} & \Includegraphics[height=8mm]{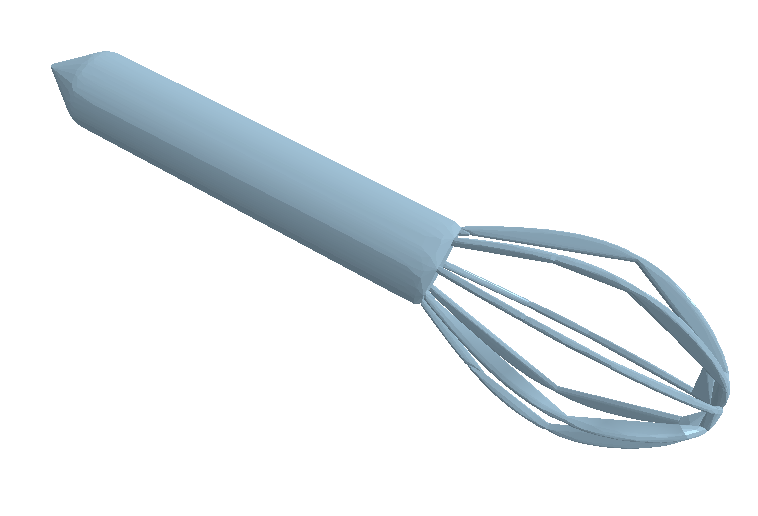} & \begin{minipage}[t]{0.5\columnwidth} \vspace{-4mm}  A whisk typically has a long, thin handle with a series of loops at the end. The loops are usually made of metal and are arranged in a spiral pattern. A whisk is a common kitchen utensil that is used to mix ingredients together or to incorporate air into a mixture. \end{minipage} & No\\
    \multirow{1}{*}{ needlenose} & \Includegraphics[height=8mm]{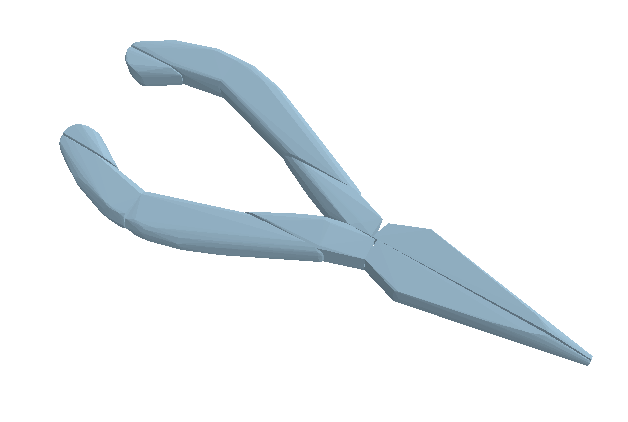} & \begin{minipage}[t]{0.5\columnwidth} \vspace{-4mm}  A needlenose plier has a long, tapered nose with a small jaw, and is used for gripping and bending wire. Needlenose pliers are a type of plier that has a long, slender nose and is used for gripping small objects and for working in tight spaces. \end{minipage} & NO \\
    \multirow{1}{*}{ plier} & \Includegraphics[height=8mm]{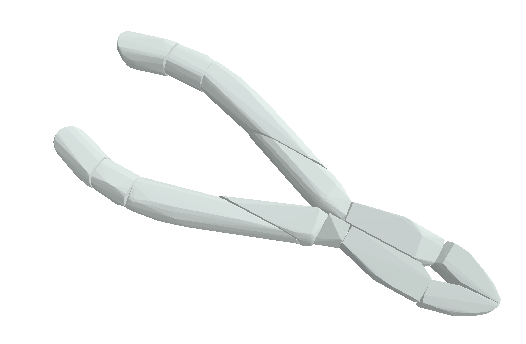} & \begin{minipage}[t]{0.5\columnwidth} \vspace{-4mm} A plier is a hand tool used for gripping objects. It consists of a pair of metal jaws with teeth that open and close when the handles are moved. Plier typically has a long, narrow neck and a tapered head that becomes progressively thinner as it extends from the neck to the tip. \end{minipage} & Yes \\
    \multirow{1}{*}{ gooseneck} & \Includegraphics[height=8mm]{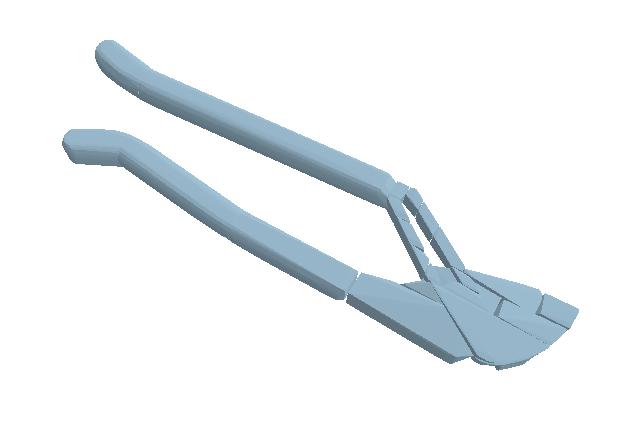} &\begin{minipage}[t]{0.5\columnwidth} \vspace{-4mm}  A gooseneck plier is a type of plier that has a long, narrow neck and a slightly curved head. The neck allows the plier to reach into tight spaces, and the curved head provides extra leverage. A gooseneck plier is commonly used to grip and bend small objects. \end{minipage} & No \\
    \multirow{1}{*}{ plier-open} & \Includegraphics[height=8mm]{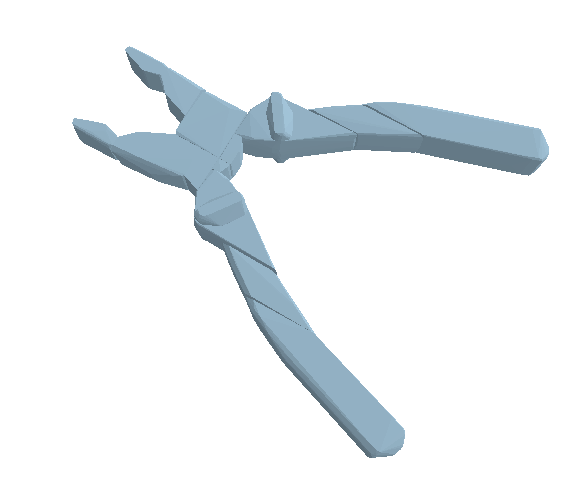} & \begin{minipage}[t]{0.5\columnwidth} \vspace{-4mm} The shape of a plier is typically long and skinny with a grip at the end. Plier is a hand tool used for various purposes such as gripping, bending and cutting. \end{minipage}  & No \\
    \multirow{1}{*}{ mallet} & \Includegraphics[height=11mm]{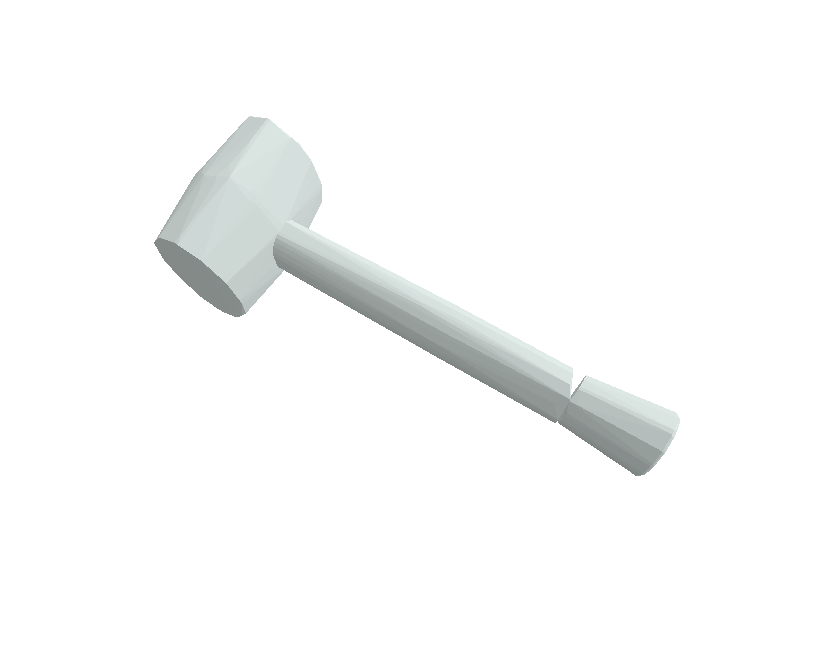} & \begin{minipage}[t]{0.5\columnwidth} \vspace{-4mm} A mallet is a tool that is used to strike another object. A mallet is a type of hammer that usually has a large head and a long handle. \end{minipage}  & No \\
    \multirow{1}{*}{ hammer} & \Includegraphics[height=7mm]{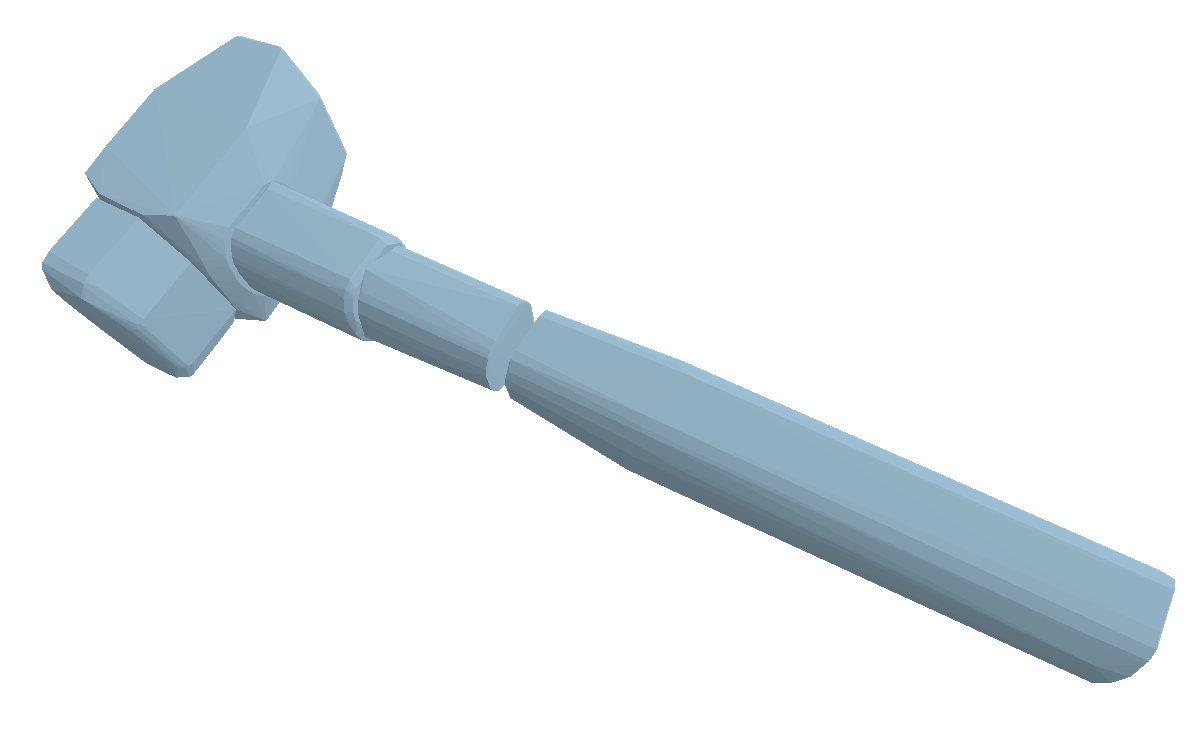} & \begin{minipage}[t]{0.5\columnwidth} \vspace{-4mm} The purpose of a hammer is to strike or hit another object. A hammer typically has a long, cylindrical handle and a heavy head. \end{minipage}  & Yes \\
    \multirow{1}{*}{ banana} & \Includegraphics[height=11mm]{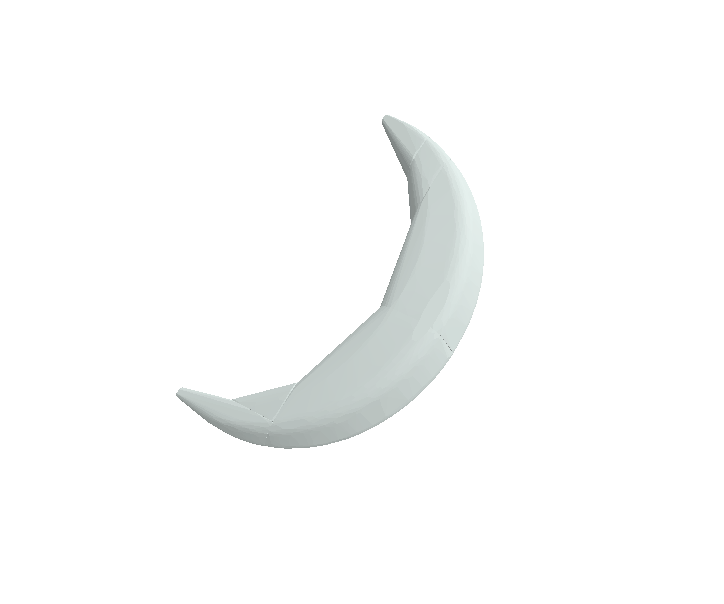} & \begin{minipage}[t]{0.5\columnwidth} \vspace{-4mm} The shape of a banana is generally long and curved, with a thin skin and fleshy inside. A banana is a curved, yellow fruit with a thick peel. \end{minipage} & No \\
    \multirow{1}{*}{ fork} & \Includegraphics[height=9mm]{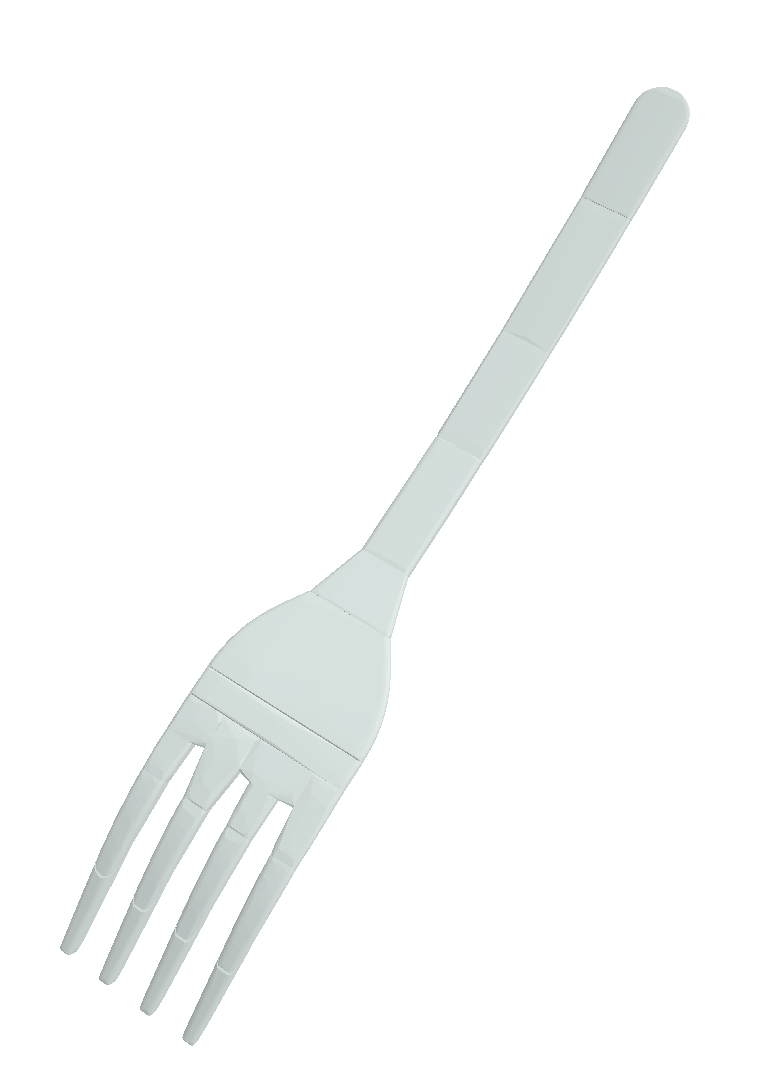} & \begin{minipage}[t]{0.5\columnwidth} \vspace{-4mm} A fork is long and thin, with three tines (prongs) at the end. A fork is a utensil that consists of a handle with several narrow tines on one end. The tines are used for piercing food and then lifting it to the mouth. \end{minipage}  & No \\
    \multirow{1}{*}{ spoon} & \Includegraphics[height=9mm]{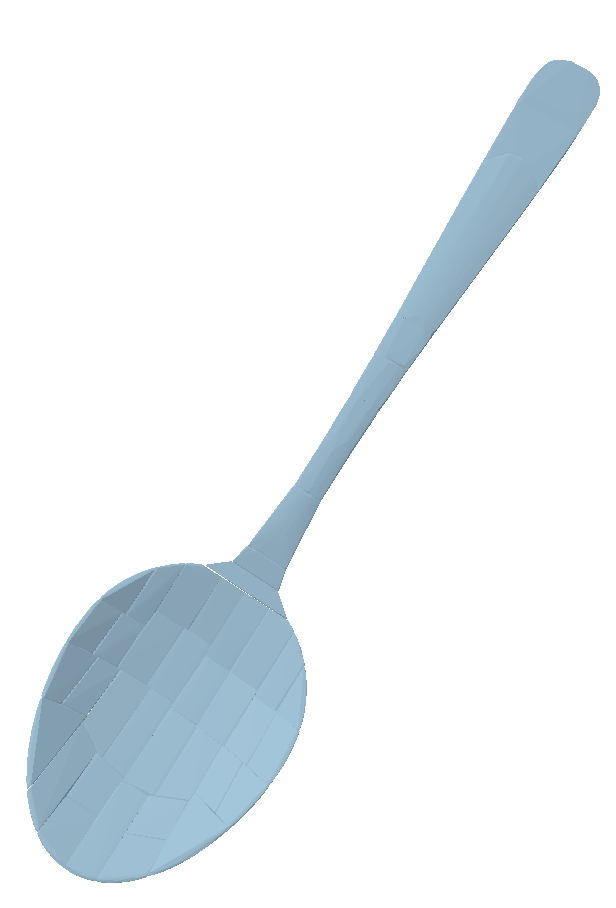} & \begin{minipage}[t]{0.5\columnwidth} \vspace{-4mm} The purpose of a spoon is to transfer a liquid or semi-solid food from a container to the mouth. A typical spoon consists of a bowl-shaped container with a handle extending from one side. The bowl is generally oval or round, and the handle generally tapers towards the end. \end{minipage} & Yes \\
    \multirow{1}{*}{ knife} & \Includegraphics[height=9mm]{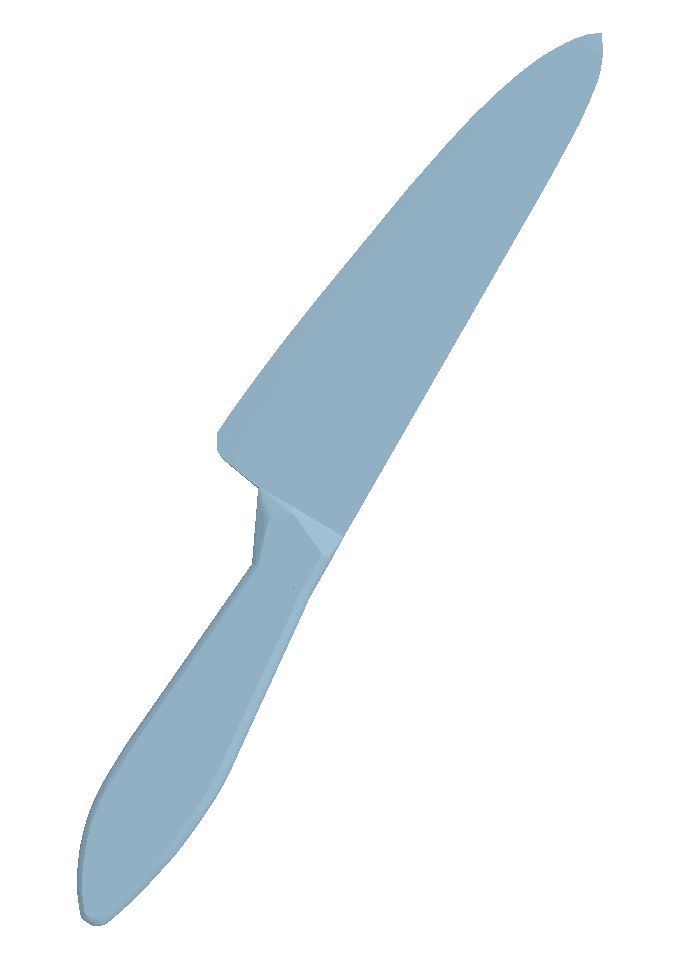} & \begin{minipage}[t]{0.5\columnwidth} \vspace{-4mm} A knife typically has a sharp, narrow blade with a pointed tip. A knife is a common kitchen utensil used for cutting and slicing food. \end{minipage} & No \\
    \multirow{1}{*}{ spatula} & \Includegraphics[height=8mm]{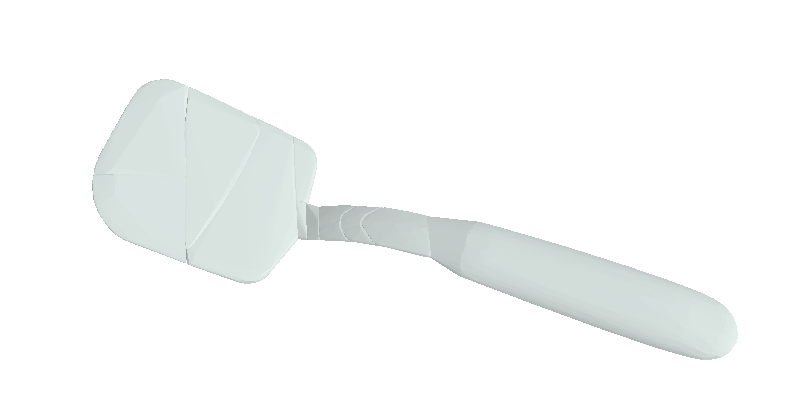} & \begin{minipage}[t]{0.5\columnwidth} \vspace{-4mm} A spatula is a kitchen utensil that is used to turn or lift food that is being cooked. It has a flat, usually slightly convex, blade that is attached to a handle. A spatula is commonly used to mix, spread, and flip food items. \end{minipage}  & No \\
    \multirow{1}{*}{ scissors} & \Includegraphics[height=10mm]{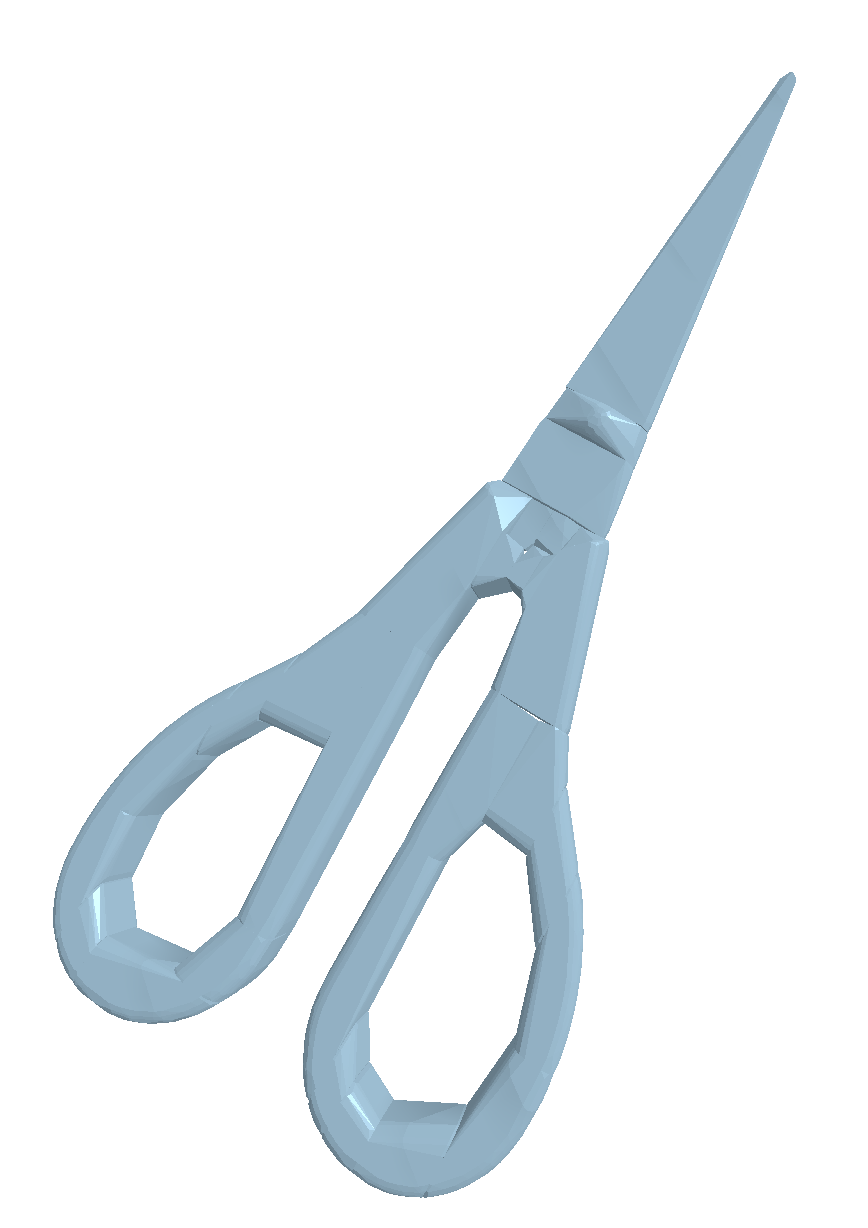} & \begin{minipage}[t]{0.5\columnwidth} \vspace{-4mm} A pair of scissors is a cutting tool that consists of two metal blades that are connected at a pivot point. A pair of scissors typically has two blades that are joined at a pivot point. \end{minipage}  & Yes \\
    \multirow{1}{*}{ wrench} & \Includegraphics[height=10mm]{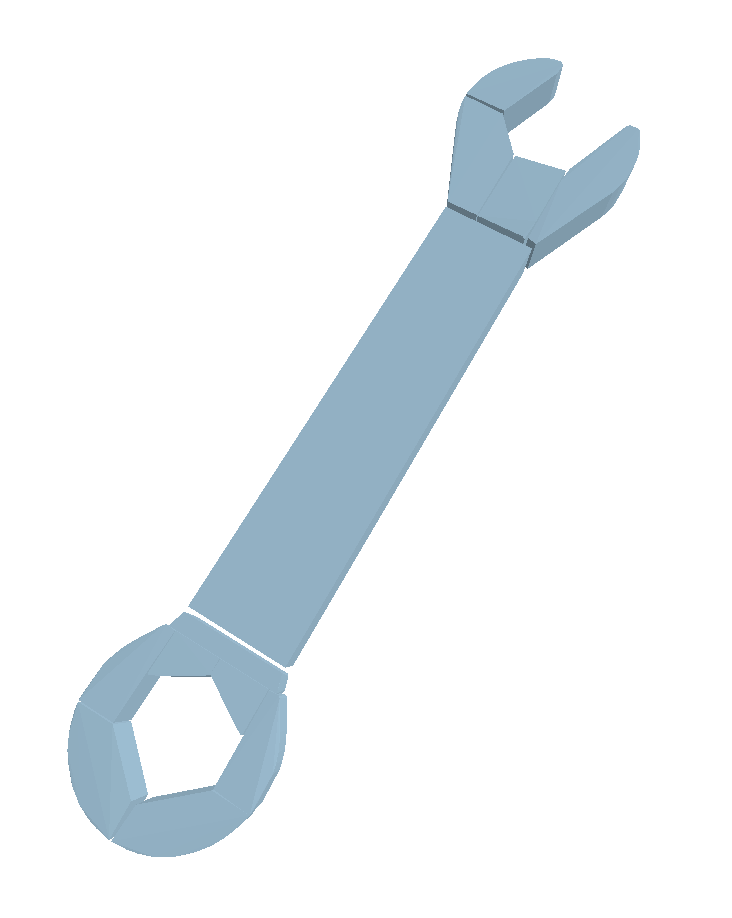} & \begin{minipage}[t]{0.5\columnwidth} \vspace{-4mm} A wrench is a tool that is used to apply torque to an object in order to loosen or tighten it. A wrench is typically long and slender with a small, metal handle. \end{minipage} & No\\
    \multirow{1}{*}{ screwdriver} & \Includegraphics[height=10mm]{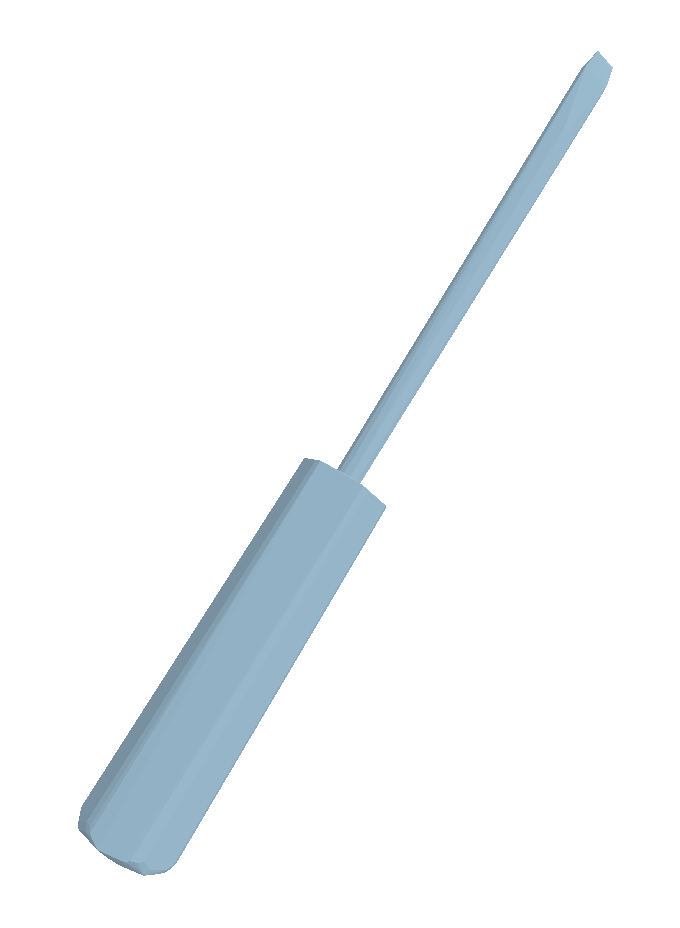} & \begin{minipage}[t]{0.5\columnwidth} \vspace{-4mm} The geometry of a screwdriver can be described as a cylindrical shape with a pointed end. A screwdriver is a tool that is used to insert and remove screws. \end{minipage} & No \\
    \multirow{1}{*}{ clamp} & \Includegraphics[height=10mm]{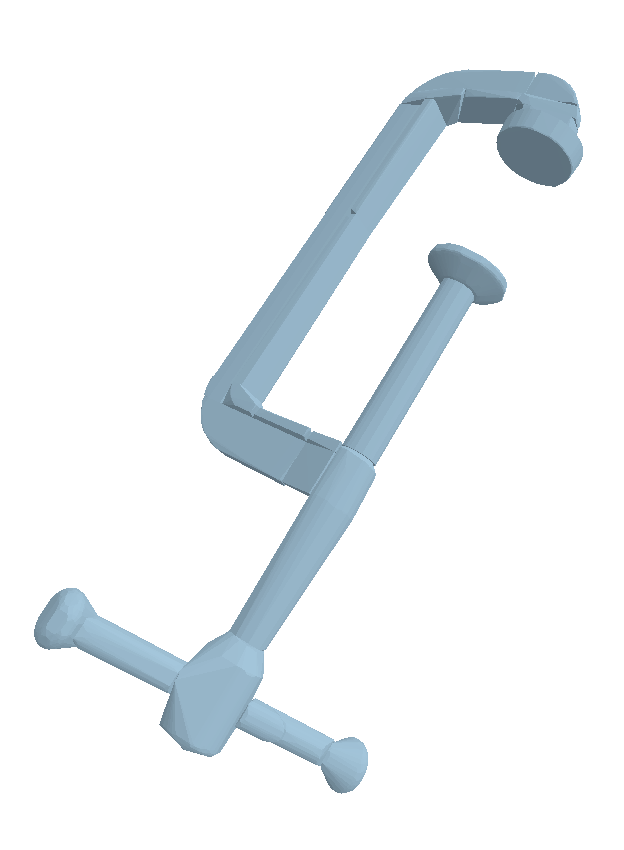} & \begin{minipage}[t]{0.5\columnwidth} \vspace{-4mm} A clamp is a mechanical device that is used to temporarily hold two or more objects together. The geometry of a clamp is typically that of a rectangular or U-shaped object with two handles. \end{minipage} & No\\
    \multirow{1}{*}{ wok} & \Includegraphics[height=9mm]{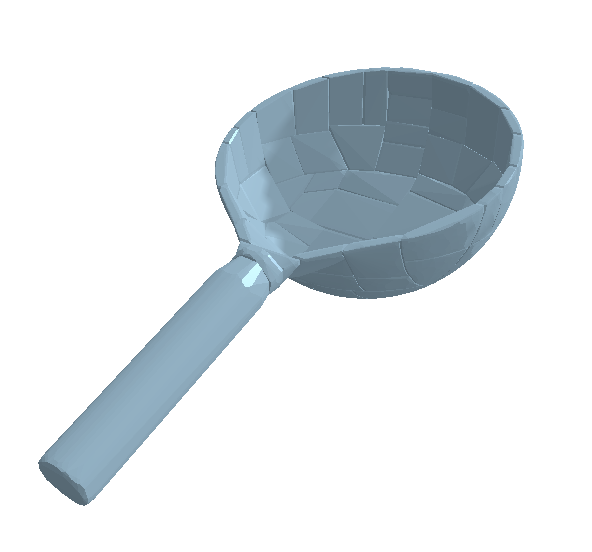} & \begin{minipage}[t]{0.5\columnwidth} \vspace{-4mm} The shape of a wok is a deep, round bowl with sloping sides. A wok is a concave-shaped cooking utensil that is most commonly used in Chinese cuisine. \end{minipage} & No \\
    \multirow{1}{*}{ pickaxe} & \Includegraphics[height=10mm]{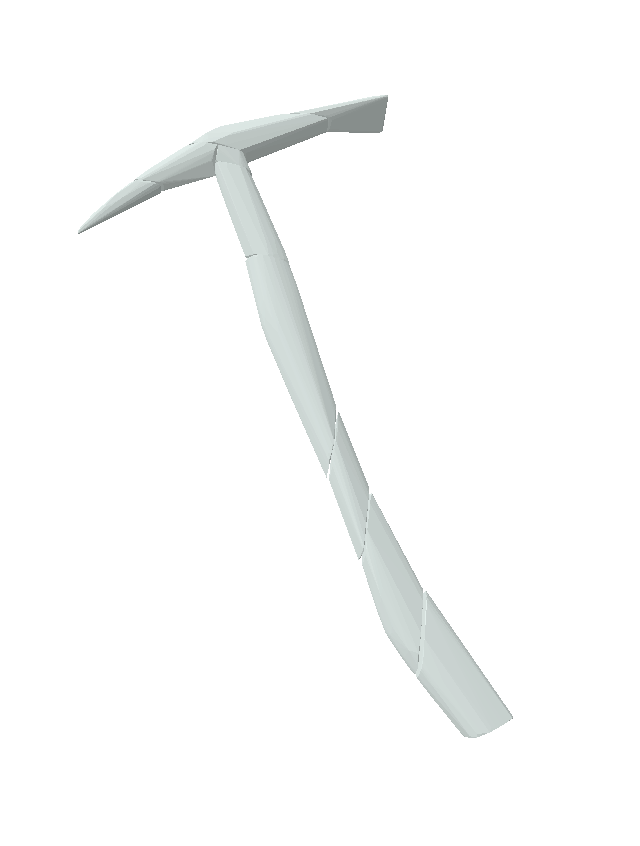} & \begin{minipage}[t]{0.5\columnwidth} \vspace{-4mm} A pickaxe is used to break up rocks and other materials. A pickaxe is a tool that has a handle attached to a head. \end{minipage} & No \\
    \multirow{1}{*}{ faucet} & \Includegraphics[height=8mm]{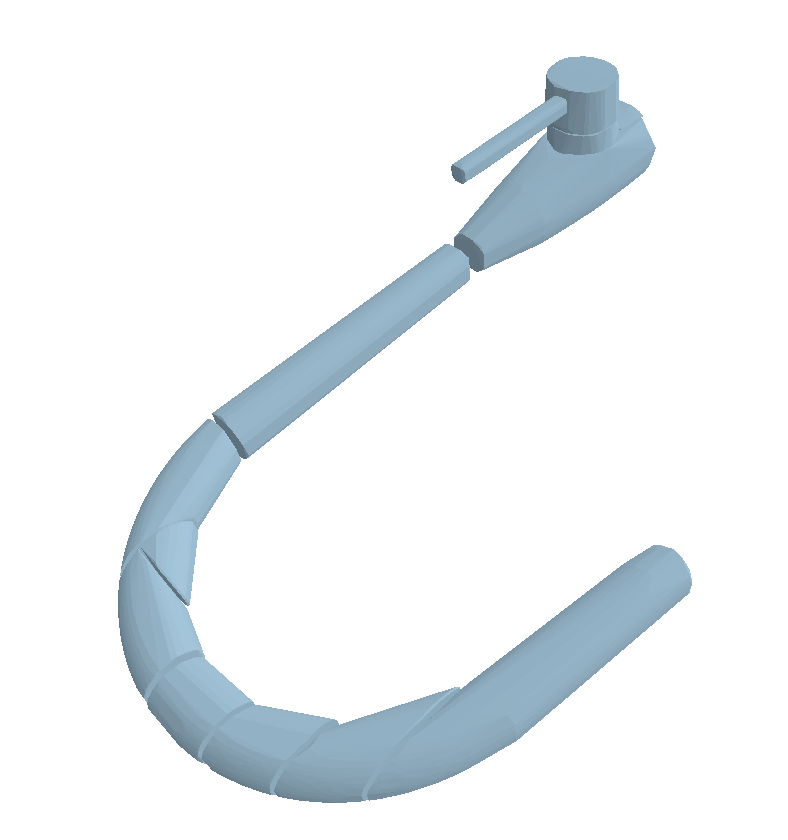} & \begin{minipage}[t]{0.5\columnwidth} \vspace{-4mm} A faucet is typically a small, thin, spout-like fixture that protrudes from a wall or sink. A faucet is a valve used to release water from a plumbing fixture, such as a sink or bathtub. \end{minipage}  & Yes \\
    \multirow{1}{*}{ dustpan} & \Includegraphics[height=8mm]{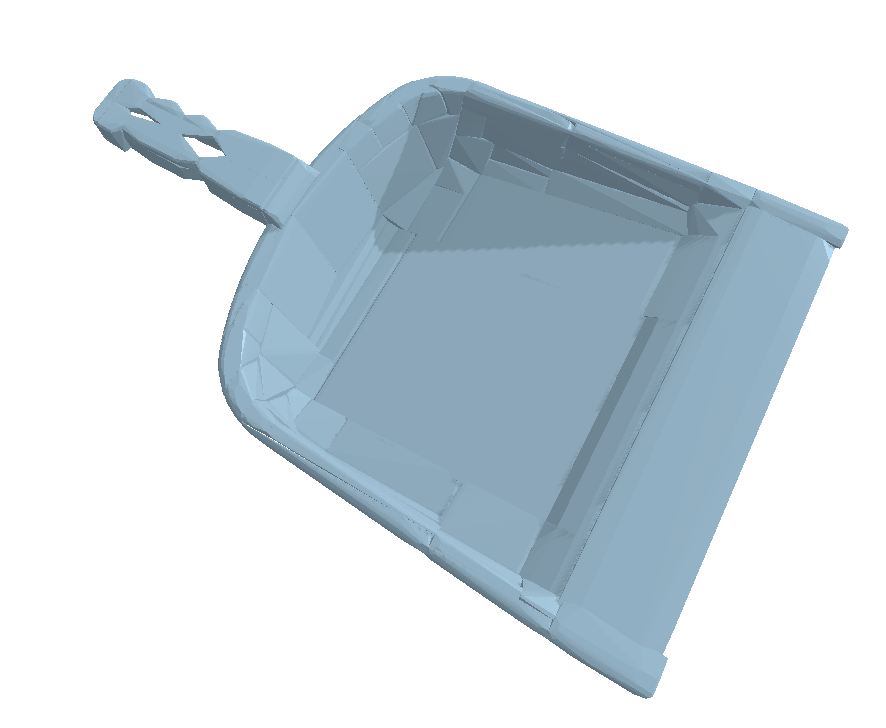} & \begin{minipage}[t]{0.5\columnwidth} \vspace{-4mm}A dustpan is a tool used for sweeping up dust and small debris from floors and other surfaces. It consists of a small, shallow pan with a handle attached to one side. A dustpan is a concave scoop with a flat bottom and flared sides. \end{minipage} & No \\
    \multirow{1}{*}{ trowel} & \Includegraphics[height=7mm]{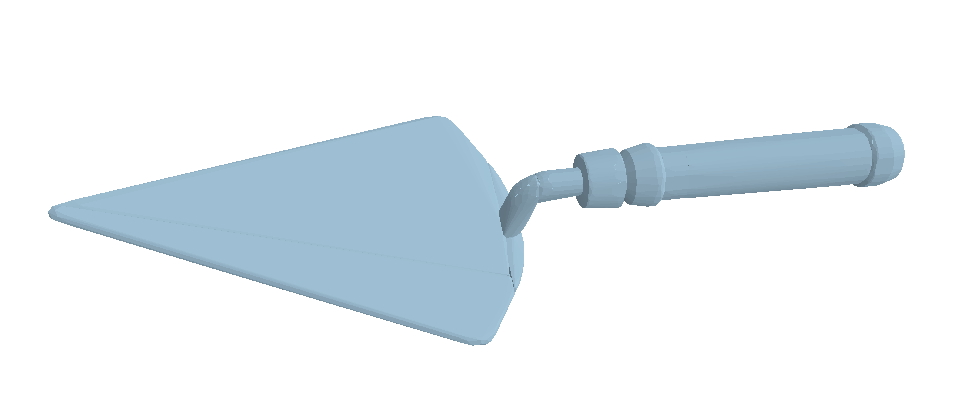} & \begin{minipage}[t]{0.5\columnwidth} \vspace{-4mm} A trowel is generally a small hand tool with a pointed, scoop-shaped blade on one end and a flat surface on the other. A trowel is a small, hand-held gardening tool with a curved, pointed blade that is used for digging, planting, and transferring small amounts of soil or other materials. \end{minipage}  & Yes\\
    \multirow{1}{*}{ ladle} & \Includegraphics[height=10mm]{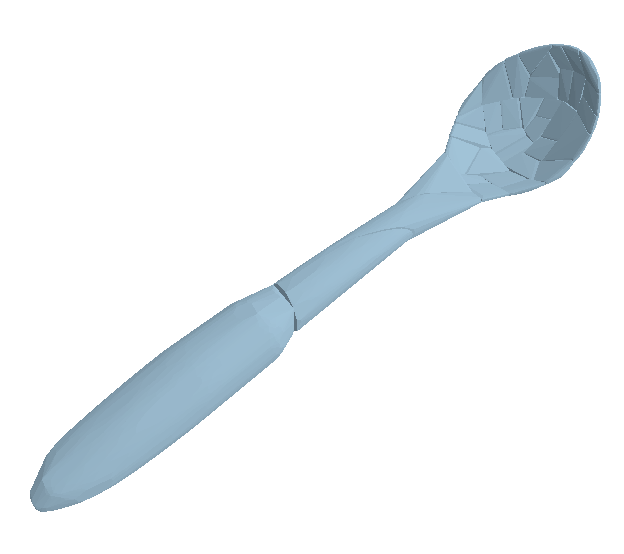} & \begin{minipage}[t]{0.5\columnwidth} \vspace{-4mm} A ladle is a tool used to transfer liquids from one container to another. A ladle typically has a long, curved handle and a large, deep, spoon-like bowl. \end{minipage} & No\\
    \multirow{1}{*}{ tongs} & \Includegraphics[height=12mm]{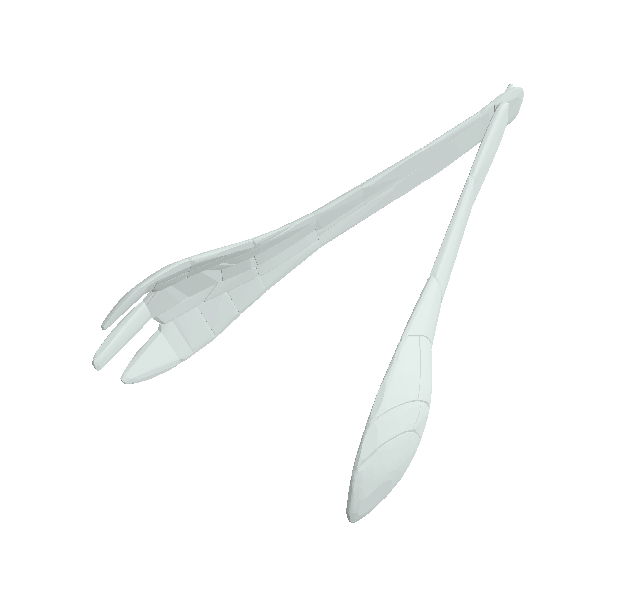} & \begin{minipage}[t]{0.5\columnwidth} \vspace{-4mm} A pair of tongs has a thin, curved metal shaft with two flat metal paddles at the end. A pair of tongs is a device used to grip and hold objects.\end{minipage}  & No \\
    \multirow{1}{*}{ gavel} & \Includegraphics[height=8mm]{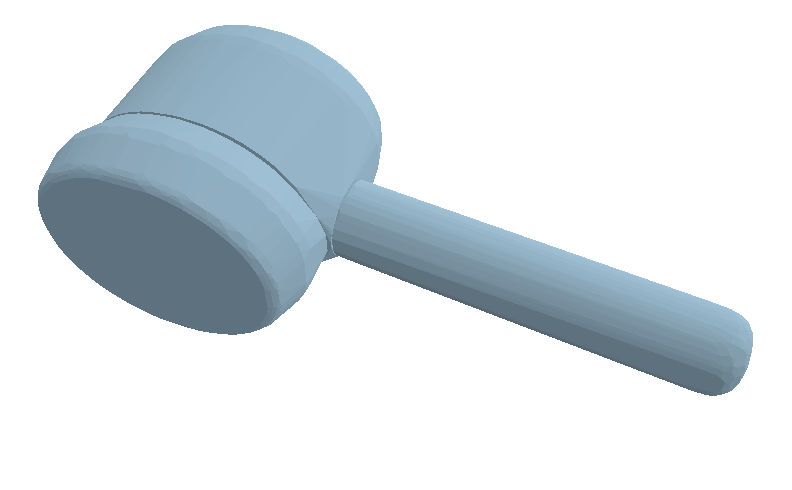} & \begin{minipage}[t]{0.5\columnwidth} \vspace{-4mm} A gavel is a small hammer that is used to strike a sound block, typically made of wood. A gavel is a mallet used to strike a block of wood, typically used by a presiding officer or auctioneer to maintain order or to signal the start and end of an auction. \end{minipage} & No \\
    \multirow{1}{*}{ squeegee} & \Includegraphics[height=8mm]{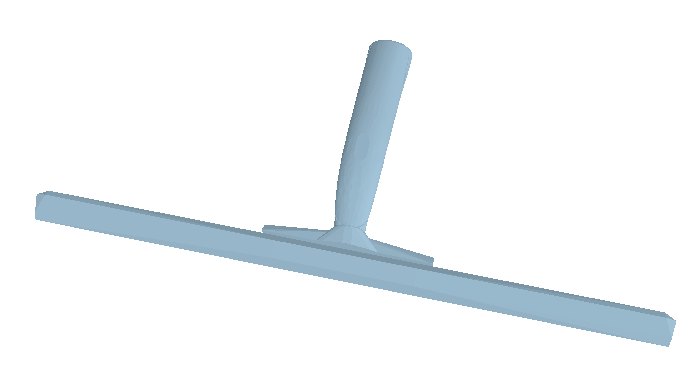} & \begin{minipage}[t]{0.5\columnwidth} \vspace{-4mm} The purpose of a squeegee is to remove water or other liquid from a surface. A squeegee is a rod-shaped tool with a flat, blunt edge, and a small handle. \end{minipage} & No \\
    \multirow{1}{*}{ powerdrill} & \Includegraphics[height=10mm]{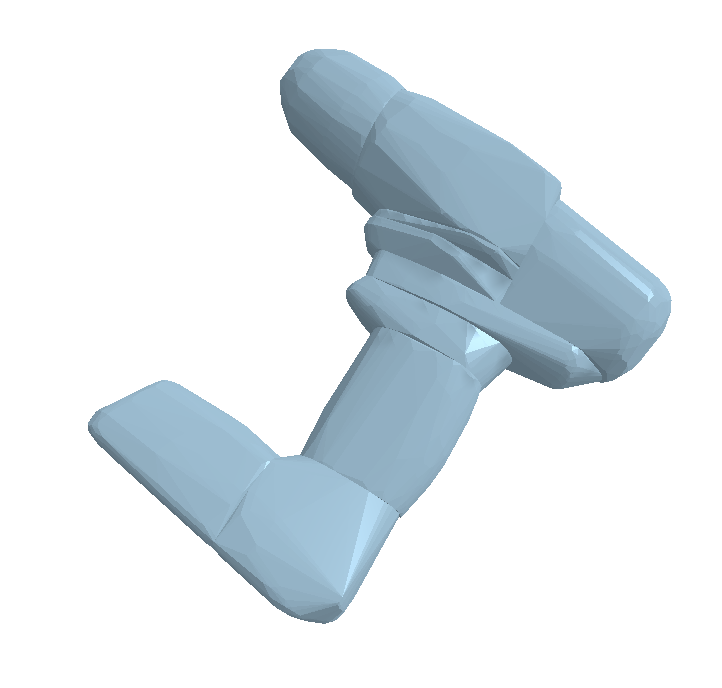} & \begin{minipage}[t]{0.5\columnwidth} \vspace{-4mm} A powerdrill is typically cylindrical in shape, with a handle attached to one side and a chuck on the other side for holding drill bits. A power drill is a tool that is used to create holes in various materials, or to fasten screws or bolts. \end{minipage} & No \\
    \multirow{1}{*}{ wineglass} & \Includegraphics[height=10mm]{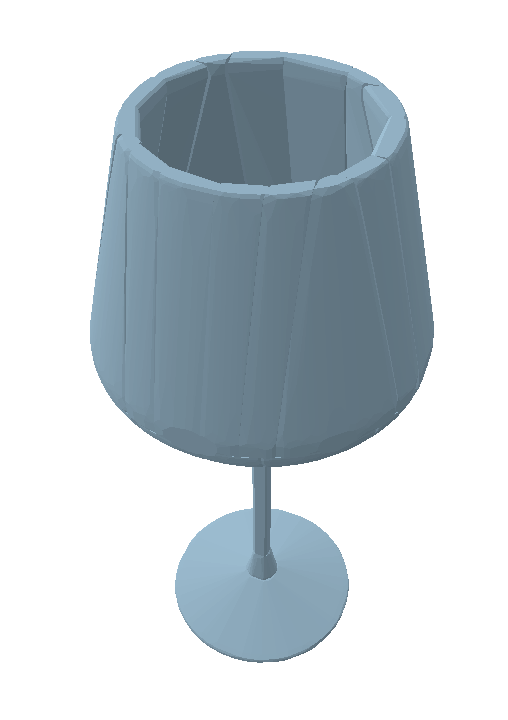} & \begin{minipage}[t]{0.5\columnwidth} \vspace{-4mm} A wineglass is a glass with a small bowl and a long stem. They are used to serve wine and are often used in restaurants. A wineglass is typically shaped with a long, thin stem and a bowl that is larger at the bottom than the top. \end{minipage} & Yes \\
    \multirow{1}{*}{ shoehorn} & \Includegraphics[height=8mm]{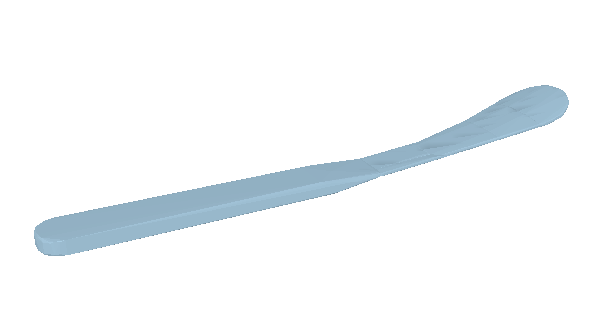} & \begin{minipage}[t]{0.5\columnwidth} \vspace{-4mm} A shoehorn is a curved, rod-shaped object used to assist in putting on shoes. A shoehorn is a curved or stepped tool designed to help slide a shoe onto the foot. \end{minipage} & No \\
    \multirow{1}{*}{ horseshoe} & \Includegraphics[height=9mm]{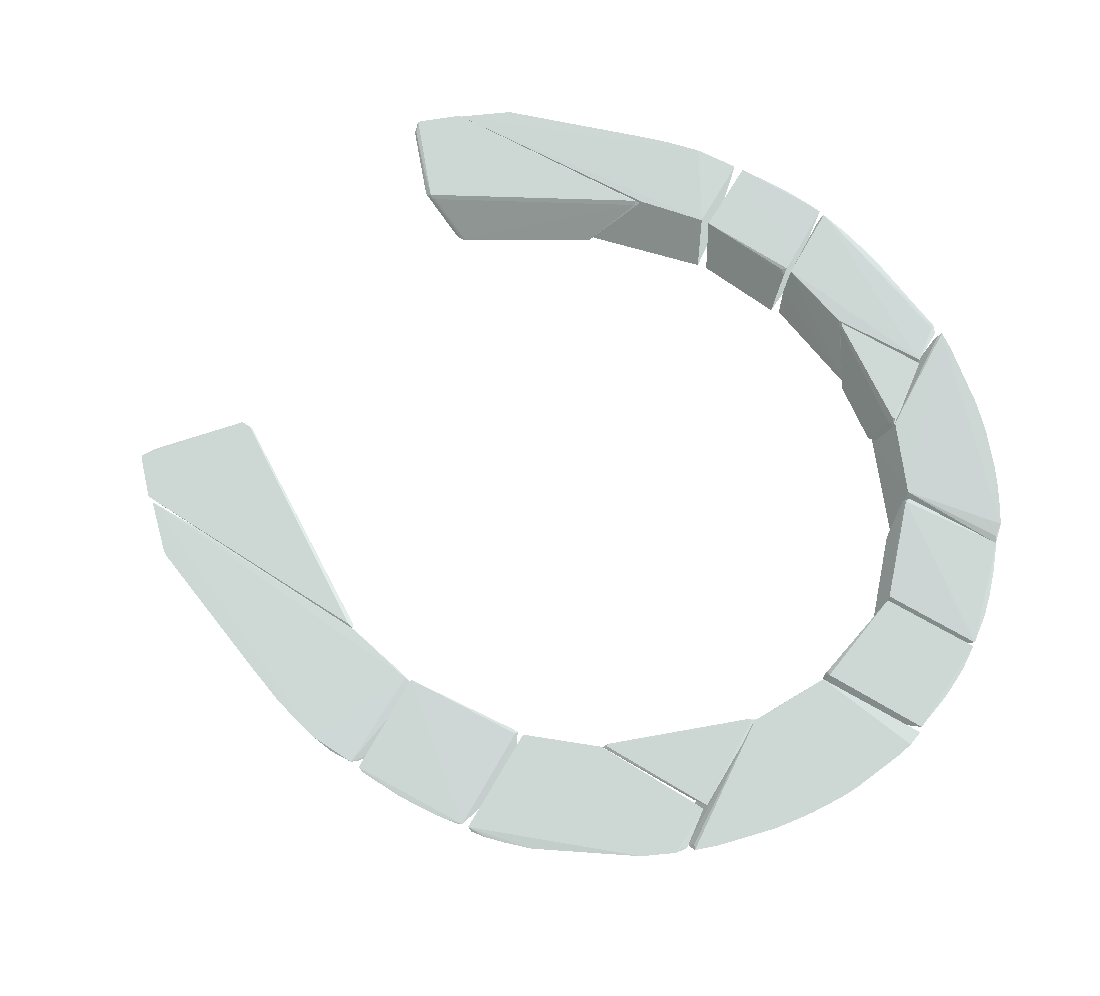} & \begin{minipage}[t]{0.5\columnwidth} \vspace{-4mm} A horseshoe is a U-shaped metal bar that is nailed to the hooves of a horse. A horseshoe is typically U-shaped, with two large curves and two smaller curves at either end. \end{minipage} & No \\
\bottomrule
% \end{tabular}}
\caption{Sample views, sample language descriptions, and the train-test split of the 36 tools considered in the paper.}
\label{app:table:tools}
% \end{table*}
\end{longtable}
% }
\end{footnotesize}

Fig.~\ref{fig:tsne} shows the t-SNE analysis of the BERT embeddings of all the tools. First we use PCA to project the 768-dimensional embeddings to 50-dimensional, and then perform t-SNE to project them to 2-dimensional for visualization.

\subsection{Task Information}
\label{app:tasks}

\begin{figure}[h]
\begin{center}
\includegraphics[width=0.8\textwidth]{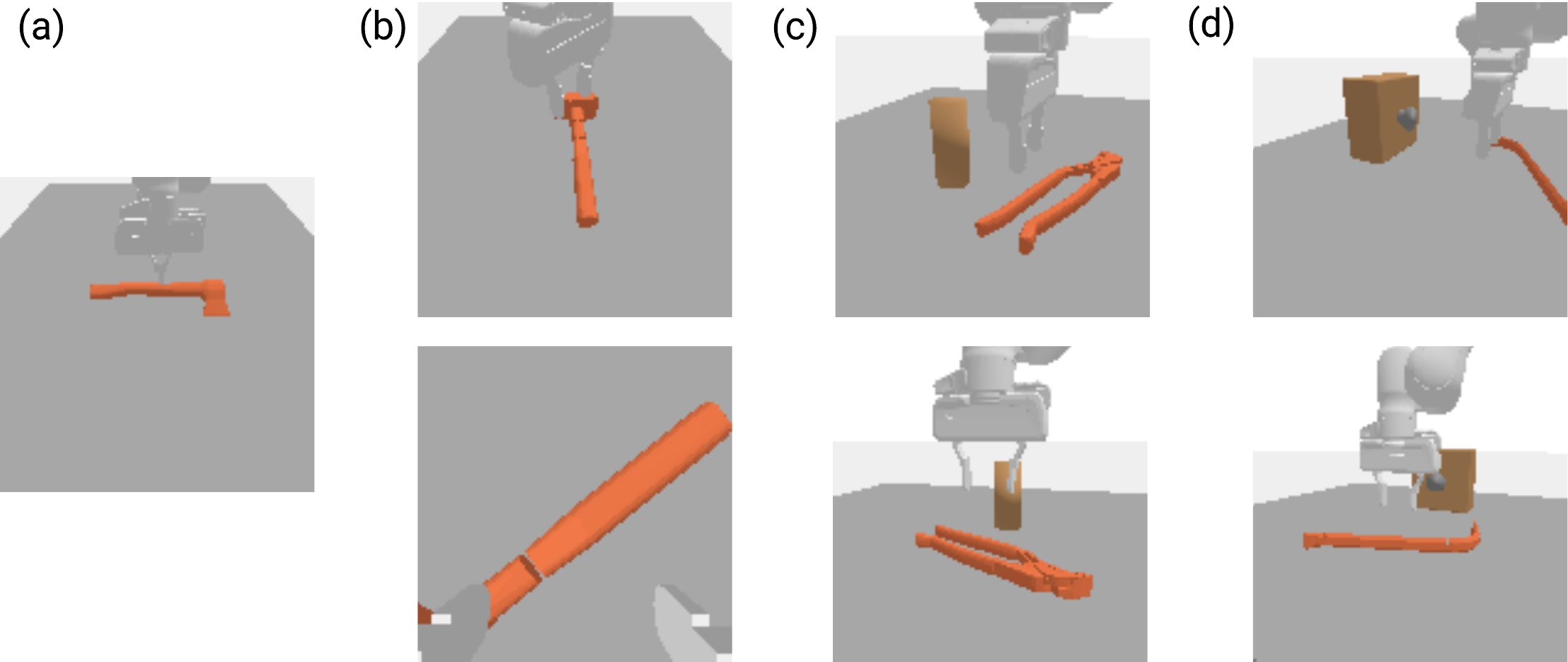}
\caption{Camera observations of the tasks: (a) pushing (single view only); (b) lifting (including a wrist view); (c) sweeping; (d) hammering.}
\label{app:fig:env-obs}
\end{center}
\end{figure}

\begin{figure}[h!]
\begin{center}
\includegraphics[width=0.9\textwidth]{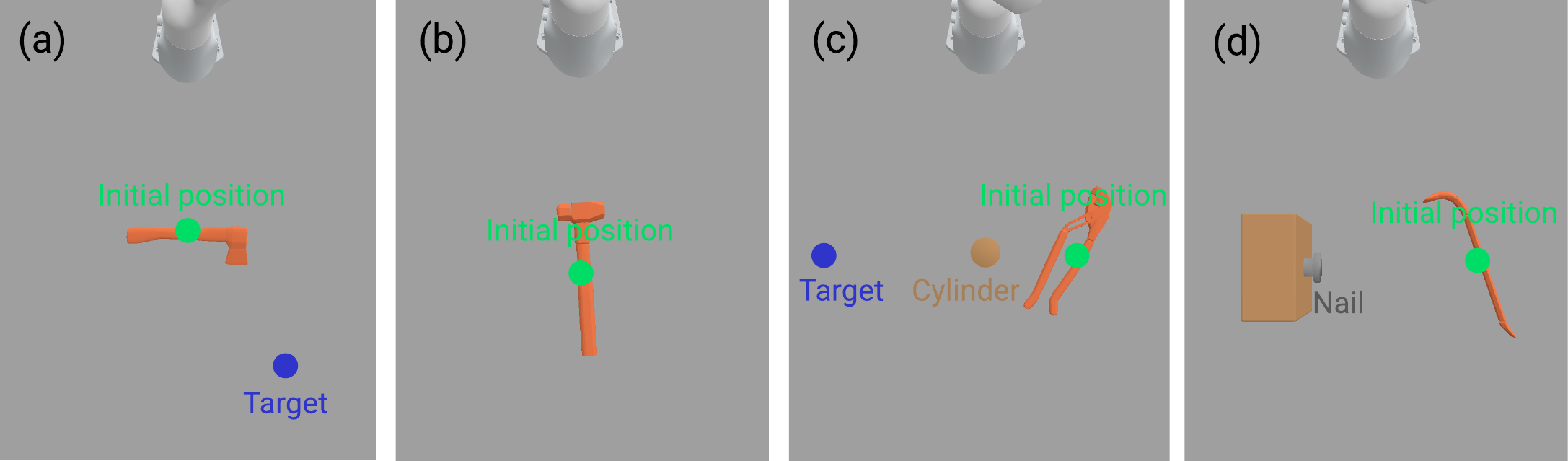}
\caption{Top-down visualization of the workspace of the tasks: (a) pushing; (b) lifting; (c) sweeping; (d) hammering.}
\label{app:fig:env-top-down}
\end{center}
\end{figure}

Table.~\ref{app:table:tasks} shows the episode length, reward function, and action space of the tasks. We find the policy can explore well in pushing and lifting tasks with relatively simple reward functions; in sweeping and hammering task, we tune the reward function carefully to guide the arm towards the cylinder/nail. Fig.~\ref{app:fig:env-obs} shows the camera observations for the four tasks. We use a single view for the pushing task as it is sufficient for the task, and dual views for other tasks. A wrist view is used in the lifting task. Fig.~\ref{app:fig:env-top-down} visualizes the workspace of the tasks including the initial position of the tools and the target.

For the hammering task, we set the lateral and torsional friction coefficient of the nail to be high ($1$ and $0.1$) in the simulator. We also make the gripper fingers longer to prevent the gripper hitting the block when attempting to hammer the nail.

Please see the included video for more visualization of the tasks.

\renewcommand{\arraystretch}{1.3} % General space between rows (1 standard)
\begin{table*}[h]
\centering
\vspace{0.0in}
\scalebox{0.85}{
\begin{tabular}{cccc}
\toprule
Task & Episode length & Reward function & Action space \\  \hline
    \multirow{1}{*}{Pushing} & 25 & 
    % \renewcommand{\arraystretch}{1}
    % \begin{tabular}{@{}c@{}} 
    $\max(0, 1-\text{distance-tool-target}$ 
    % \\ Target at $[x=0.75, y=0.15]$ \end{tabular} 
    &
    \begin{tabular}{@{}c@{}}$[-0.05, 0.15]m/s$ in $x$ \\ $[-0.1, 0.1]m/s$ in $y$ \\ $[-\pi/4, \pi/4]\text{rad}/s$ in yaw \end{tabular} \\ \hline
    \multirow{1}{*}{Lifting} & 25 & 
    \begin{tabular}{@{}c@{}} $0.1*\max(0, 1-\text{distance-EE-tool}) +$ \\ $0.5*\max(0, 1-\text{distance-tool-target})$ 
    % Target at $[z=0.1]$ 
    \end{tabular} &
    \begin{tabular}{@{}c@{}}$[-0.1, 0.1]m/s$ in $x$, $y$, and $z$ \\ $[-\pi/4, \pi/4]\text{rad}/s$ in yaw \end{tabular} \\ \hline
    \multirow{1}{*}{Sweeping} & 40 & 
    \begin{tabular}{@{}c@{}} $0.1*\max(0, 1-\text{distance-EE-tool}) +$ \\ $0.1*\max(0, 1-\text{distance-tool-cylinder}) +$ \\ $0.5*\max(0, 1-\text{distance-cylinder-target})$
    % Target at $[z=0.1]$ 
    \end{tabular} &
    \begin{tabular}{@{}c@{}}$[-0.2, 0.2]m/s$ in $x$, $y$, and $z$ \\ $[-\pi/4, \pi/4]\text{rad}/s$ in yaw \end{tabular} \\ \hline
    \multirow{1}{*}{Hammering} & 40 & 
    \begin{tabular}{@{}c@{}} $0.1*\max(0, 1-\text{distance-EE-tool}) +$ \\ $0.1*\max(0, 1-\text{distance-tool-nail}) +$ \\ $0.5*\max(0, 1-\text{distance-nail-hole\textunderscore end})$ \end{tabular} &
    \begin{tabular}{@{}c@{}}$[-0.2, 0.2]m/s$ in $x$, $y$, and $z$ \\ $[-\pi/4, \pi/4]\text{rad}/s$ in yaw \end{tabular} \\
\bottomrule
\end{tabular}
}
\caption{Episode length, reward function, and action space for the four tasks. Distance-\{A\}-\{B\} denotes distance from A to B, normalized by the initial distance. EE denotes end-effector of the arm. See Fig.~\ref{app:fig:env-top-down} for visualization of the task space and target.}
\label{app:table:tasks}
\end{table*}

% For all four tasks, we show the undiscounted cumulative reward over the episode horizon.

\newpage
\subsection{Model Architecture}
\label{app:model}

For all policies, the image encoder $\pi_o$ contains three convolutional layers for either view of the image input; the three layers have kernel size $[7\times7, 5\times5, 3\times3]$, stride size $[4,3,2]$, no padding, and channel size $[4, 8, 16]$. The language head $\pi_l$ contains a single fully-connected layers with 128-dimensional output. Both the actor head $\pi_a$ and critic head $\pi_c$ have two hidden layers of size 128 (except for AT-XL with hidden size 256). All convolutional layers and fully-connected layers are followed with a ReLU activation. The first layer in $\pi_a$ and $\pi_c$ are additionally normalized by Layernorm \cite{ba2016layer}. 
% See Fig.~\ref{fig:model} for visualization.

\subsection{Training Hyper-parameters}
\label{app:training}
The hyper-parameters used for meta-learning (shared among AT-LA, AT-TinyLA, AT, AT-XL) and vanilla multi-task learning (shared between SAC-LA, SAC) are outlined in Table~\ref{app:table:hyper}. We ensure meta-learning and multi-task learning sample the same amount of transitions from environments.

\vspace{5mm}
\renewcommand{\arraystretch}{1.1} % General space between rows (1 standard)
\begin{table*}[h]
\centering
\scalebox{0.9}{
\begin{tabular}{ccc}
    \toprule
    Setting & Meta-learning & Multi-task learning \\ \midrule
    $\#$ training steps & 1000 (iteration) & \begin{tabular}{@{}c@{}} 2.5e6 (pushing/lifting) \\ 4e6 (sweeping/hammering) \end{tabular} \\
    Meta replay buffer size & 30000 & --- \\
    Base replay buffer size & $\infty$ & 100000 \\
    Replay ratio & \multicolumn{2}{c}{16} \\
    $N$ & 1 & --- \\
    $M$ & 2 & --- \\
    $B$ & 5 & --- \\
    $B_\nu$ & \multicolumn{2}{c}{10} \\ \midrule
    \textbf{Optimization} \\
    Optimizer & \multicolumn{2}{c}{Adam} \\
    Batch size & \multicolumn{2}{c}{128} \\
    Discount factor & \multicolumn{2}{c}{0.99} \\
    SAC entropy coefficient & \multicolumn{2}{c}{0.01} \\
    SAC actor update period & \multicolumn{2}{c}{1} \\
    Base learning rate & \multicolumn{2}{c}{3e-4} \\
    Meta learning rate & 1e-3 & --- \\ \midrule
    \textbf{Hardware Resource} \\
    $\#$ CPU threads & \multicolumn{2}{c}{20} \\
    GPU & \multicolumn{2}{c}{Nvidia RTX 2080Ti} \\
    $\#$ hours for runtime & \multicolumn{2}{c}{6 (pushing/lifting), 10 (sweeping), 16 (hammering)} \\ 
    \bottomrule
\end{tabular}}
\caption{Hyper-parameters used for meta-learning and multi-task learning.}
\label{app:table:hyper}
\end{table*}

\newpage
\subsection{Results}
\label{app:results}

Table \ref{tab:results} below shows the results in Fig.~\ref{fig:results} in numbers. 

\vspace{5mm}
\setlength{\tabcolsep}{4pt} % Make spacing between table columns smaller; default value: 6pt
\renewcommand{\arraystretch}{1.1} % General space between rows (1 standard)
\begin{table}[h]
\scriptsize
\begin{center}
\begin{tabular}{cccccccccc}
    \toprule
    & \multicolumn{8}{c}{\textbf{Pushing}}\\
    \cmidrule{2-10}
    {Method} & Crowbar & Paint Roller & Plier & Hammer & Spoon & Scissors & Faucet & Trowel & Wineglass\\ \midrule
    ATLA & \bftab 15.4 $\pm$ 0.6 & \bftab 19.9 $\pm$ 0.2 & 19.9 $\pm$ 0.4 & \bftab 11.1 $\pm$ 0.1 & 10.0 $\pm$ 0.3 & \bftab 18.2 $\pm$ 0.5 & \bftab 14.4 $\pm$ 0.5 & \bftab 16.9 $\pm$ 2.2 & 4.5 $\pm$ 0.3 \\
    AT-TinyLA & 12.3 $\pm$ 1.8 & 19.2 $\pm$ 0.9 & \bftab 20.4 $\pm$ 0.2 & 10.7 $\pm$ 0.3 & 10.5 $\pm$ 0.8 & 17.0 $\pm$ 0.8 & 8.8 $\pm$ 0.9 & 12.5 $\pm$ 1.1 & 7.4 $\pm$ 1.6 \\
    AT & 7.8 $\pm$ 0.3 & 17.4 $\pm$ 1.0 & 18.4 $\pm$ 0.5 & 10.0 $\pm$ 0.4 & 9.4 $\pm$ 0.2 & 15.6 $\pm$ 0.4 & 9.6 $\pm$ 1.2 & 15.8 $\pm$ 1.5 & 6.1 $\pm$ 0.9 \\
    AT-XL & 12.2 $\pm$ 2.5 & 19.0 $\pm$ 0.8 & 19.6 $\pm$ 0.5 & 10.8 $\pm$ 0.1 & \bftab 11.3 $\pm$ 0.2 & 16.3 $\pm$ 1.4 & 14.1 $\pm$ 1.8 & 15.4 $\pm$ 0.7 & 7.1 $\pm$ 0.6 \\
    SAC-LA & 13.4 $\pm$ 0.8 & 16.8 $\pm$ 0.6 & 18.4 $\pm$ 0.5 & 10.7 $\pm$ 1.0 & 7.8 $\pm$ 1.9 & 14.2 $\pm$ 1.3 & 10.1 $\pm$ 1.6 & 12.7 $\pm$ 2.5 & \bftab 7.8 $\pm$ 1.1 \\
    SAC & 12.9 $\pm$ 1.1 & 12.2 $\pm$ 4.7 & 16.9 $\pm$ 1.5 & 10.6 $\pm$ 0.6 & 9.2 $\pm$ 1.5 & 12.0 $\pm$ 5.6 & 10.4 $\pm$ 3.8 & 11.2 $\pm$ 3.3 & 6.4 $\pm$ 0.2 \\
    \bottomrule
    & \multicolumn{8}{c}{\textbf{Lifting}} \\
    \midrule
    ATLA & \bftab 7.4 $\pm$ 0.8 & \bftab 6.5 $\pm$ 0.8 & \bftab 9.0 $\pm$ 0.9 & 9.6 $\pm$ 1.3 & \bftab 5.5 $\pm$ 1.0 & \bftab 2.1 $\pm$ 0.0 & \bftab 3.5 $\pm$ 0.5 & \bftab 5.0 $\pm$ 0.5 & 2.3 $\pm$ 0.6 \\
    AT-TinyLA & 2.6 $\pm$ 0.3 & 5.7 $\pm$ 0.8 & 9.0 $\pm$ 1.2 & \bftab 10.4 $\pm$ 0.3 & 5.0 $\pm$ 0.8 & 2.0 $\pm$ 0.4 & 2.3 $\pm$ 0.4 & 1.9 $\pm$ 0.2 & 2.4 $\pm$ 0.5 \\
    AT & 5.0 $\pm$ 0.5 & 4.6 $\pm$ 0.8 & 7.7 $\pm$ 1.3 & 9.1 $\pm$ 2.3 & 3.7 $\pm$ 0.8 & 2.0 $\pm$ 0.3 & 1.8 $\pm$ 0.2 & 3.4 $\pm$ 0.9 & \bftab 2.5 $\pm$ 0.6 \\
    AT-XL & 6.8 $\pm$ 1.0 & 5.7 $\pm$ 1.0 & 8.5 $\pm$ 0.3 & 9.0 $\pm$ 0.4 & 4.9 $\pm$ 0.8 & 1.8 $\pm$ 0.0 & 2.4 $\pm$ 0.2 & 4.2 $\pm$ 1.2 & 2.9 $\pm$ 0.7 \\
    SAC-LA & 5.1 $\pm$ 0.1 & 4.4 $\pm$ 0.3 & 7.7 $\pm$ 0.2 & 8.3 $\pm$ 0.1 & 4.9 $\pm$ 0.1 & 1.8 $\pm$ 0.0 & 3.0 $\pm$ 0.1 & 3.8 $\pm$ 0.1 & 1.8 $\pm$ 0.1 \\
    SAC & 6.6 $\pm$ 1.5 & 4.9 $\pm$ 1.1 & 6.9 $\pm$ 1.4 & 7.9 $\pm$ 1.5 & 4.2 $\pm$ 0.6 & 1.6 $\pm$ 0.6 & 3.0 $\pm$ 0.4 & 3.6 $\pm$ 0.6 & 2.2 $\pm$ 0.6 \\
    \bottomrule
    %\\
    & \multicolumn{8}{c}{\textbf{Sweeping}} \\
    \midrule
    ATLA & \bftab 25.5 $\pm$ 1.0 & 5.2 $\pm$ 2.4 & \bftab 18.4 $\pm$ 3.5 & \bftab 26.8 $\pm$ 2.6 & \bftab 22.5 $\pm$ 2.7 & \bftab 21.7 $\pm$ 2.0 & \bftab 20.9 $\pm$ 1.6 & \bftab 24.3 $\pm$ 2.2 & \bftab 14.4 $\pm$ 1.8 \\
    AT-TinyLA & 21.2 $\pm$ 0.7 & 4.6 $\pm$ 0.2 & 16.5 $\pm$ 0.6 & 22.3 $\pm$ 0.8 & 18.0 $\pm$ 0.1 & 19.2 $\pm$ 1.0 & 15.2 $\pm$ 0.8 & 21.2 $\pm$ 0.6 & 12.5 $\pm$ 0.4 \\
    AT & 20.0 $\pm$ 0.8 & 7.7 $\pm$ 3.7 & 17.4 $\pm$ 3.5 & 19.3 $\pm$ 1.5 & 18.4 $\pm$ 1.3 & 18.0 $\pm$ 3.0 & 18.0 $\pm$ 3.2 & 14.9 $\pm$ 2.9 & 12.0 $\pm$ 2.2 \\
    AT-XL & 19.0 $\pm$ 3.0 & 6.1 $\pm$ 1.4 & 18.2 $\pm$ 2.8 & 21.0 $\pm$ 2.8 & 17.1 $\pm$ 2.4 & 16.8 $\pm$ 3.7 & 16.3 $\pm$ 3.4 & 15.1 $\pm$ 8.4 & 7.6 $\pm$ 3.6 \\
    SAC-LA & 18.4 $\pm$ 4.0 & \bftab 9.0 $\pm$ 3.7 & 15.5 $\pm$ 3.2 & 20.7 $\pm$ 2.2 & 17.4 $\pm$ 3.0 & 18.4 $\pm$ 3.7 & 16.7 $\pm$ 2.9 & 18.6 $\pm$ 1.8 & 12.2 $\pm$ 1.1 \\
    SAC & 17.6 $\pm$ 2.5 & 7.7 $\pm$ 2.1 & 12.4 $\pm$ 2.4 & 12.3 $\pm$ 4.7 & 17.3 $\pm$ 5.4 & 17.8 $\pm$ 2.4 & 15.9 $\pm$ 2.8 & 17.1 $\pm$ 0.2 & 10.6 $\pm$ 3.3 \\
    \bottomrule
    %\\
    & \multicolumn{8}{c}{\textbf{Hammering}} \\
    \midrule
    ATLA & \bftab 12.9 $\pm$ 1.0 & \bftab 12.5 $\pm$ 1.3 &  \bftab 10.9 $\pm$ 1.7 & 14.4 $\pm$ 0.9 & \bftab 12.4 $\pm$ 1.3 & 3.9 $\pm$ 0.3 & 3.8 $\pm$ 0.6 & \bftab 11.7 $\pm$ 1.2 & 10.4 $\pm$ 1.5 \\
    AT-TinyLA & 12.1 $\pm$ 1.9 & 12.0 $\pm$ 2.3 & 10.8 $\pm$ 1.5 &  \bftab 16.8 $\pm$ 1.0 & 11.5 $\pm$ 1.4 & \bftab 4.4 $\pm$ 0.2 &  \bftab 4.1 $\pm$ 0.8 & 10.3 $\pm$ 0.8 & \bftab 11.2 $\pm$ 0.8 \\
    AT & 6.0 $\pm$ 0.9 & 9.7 $\pm$ 2.0 & 10.3 $\pm$ 2.6 & 12.4 $\pm$ 1.9 & 7.2 $\pm$ 1.7 & 3.7 $\pm$ 0.1 & 3.1 $\pm$ 0.1 & 9.0 $\pm$ 2.8 & 7.1 $\pm$ 2.0 \\
    AT-XL & 6.5 $\pm$ 1.2 & 8.3 $\pm$ 1.5 & 11.3 $\pm$ 1.6 & 12.1 $\pm$ 1.7 & 7.2 $\pm$ 1.6 & 3.7 $\pm$ 0.1 & 3.1 $\pm$ 0.3 & 10.4 $\pm$ 3.4 & 8.2 $\pm$ 1.7 \\
    SAC-LA & 4.3 $\pm$ 1.1 & 3.8 $\pm$ 1.3 & 9.0 $\pm$ 3.2 & 9.3 $\pm$ 2.0 & 7.8 $\pm$ 1.6 & 3.2 $\pm$ 0.5 & 3.3 $\pm$ 0.3 & 4.7 $\pm$ 1.3 & 3.0 $\pm$ 0.7 \\
    SAC & 5.3 $\pm$ 3.1 & 5.1 $\pm$ 3.2 & 6.8 $\pm$ 4.2 & 7.4 $\pm$ 5.3 & 3.4 $\pm$ 1.1 & 3.0 $\pm$ 0.6 & 3.2 $\pm$ 0.6 & 3.5 $\pm$ 0.4 & 3.2 $\pm$ 1.0 \\
    \bottomrule
\end{tabular}
\small
\vspace{5pt}
\caption{Post-adaptation reward in mean and standard deviation over 3 seeds across 4 tasks and 9 test tools.}
\label{tab:results}
\end{center}
\end{table}

% \putbib[bib-lang]  % .bib
% \end{bibunit}

\end{document}